\def\year{2016}\relax
\documentclass[journal,transmag]{IEEEtran}

\usepackage{times}
\usepackage{helvet}
\usepackage{courier}

\usepackage{graphicx}
\usepackage{amsmath,amssymb} 
\usepackage{color}
\usepackage{times}
\usepackage{epsfig}
\usepackage{float}
\usepackage{url}
\usepackage{subfig}
\usepackage{multirow}

\usepackage{array}
\newcolumntype{x}[1]{>{\centering\arraybackslash\hspace{0pt}}m{#1}}
\usepackage{epstopdf}
\def\footnoterule{\relax%
	\kern-5pt
	\hbox to \columnwidth{\hfill\vrule width \columnwidth height 0.4pt\hfill}
	\kern4.6pt
	}
\makeatother

\newcommand{\cc}{\textcolor{black}}

\newcommand{\KG}{\textcolor{black}}

\newcommand{\iccvf}{\textcolor{black}}

\newcommand{\BX}{\textcolor{black}}
\begin{document}
%
\title{Pixel Objectness} 

\author{
	\IEEEauthorblockN{Suyog~Dutt~Jain and Bo~Xiong and
		Kristen~Grauman}
	\IEEEauthorblockA{Department of Computer Science \\ The University of Texas at Austin \\ \url{http://vision.cs.utexas.edu/projects/pixelobjectness/}}
}

\maketitle

\begin{abstract}
	We propose an end-to-end learning framework for foreground object segmentation.  Given a single novel image, our approach produces a pixel-level mask for all ``object-like" regions---even for object categories never seen during training.  We formulate the task as a structured prediction problem of assigning a foreground/background label to each pixel, implemented using a deep fully convolutional network.  Key to our idea is training with a mix of \emph{image-level} object category examples together with relatively few images with \emph{boundary-level} annotations.  \iccvf{Our method substantially improves the state-of-the-art on foreground segmentation for ImageNet and MIT Object Discovery datasets.}  Furthermore, on over 1 million images, we show that it generalizes well to segment object categories unseen in the foreground maps used for training.   Finally, we demonstrate how our approach benefits image retrieval and image retargeting, both of which flourish when given our high-quality foreground maps.
\end{abstract}

\section{Introduction}\label{sec:introduction}

Foreground object segmentation is a fundamental vision problem with several applications. For example, a visual search system can use foreground segmentation to focus on the important objects in the query image, ignoring background clutter.  It is also a prerequisite in graphics applications like rotoscoping and image retargeting.  Knowing the spatial extent of objects can also benefit downstream vision tasks like scene understanding, caption generation, and summarization. In any such setting, it is crucial to segment ``generic" objects in a \emph{category-independent} manner.  That is, the system must be able to identify object boundaries for objects it has never encountered during training.\footnote{This differentiates the problem from traditional recognition or ``semantic segmentation"~\cite{noh2015learning,crfasrnn_iccv2015,long_shelhamer_fcn,chen14semantic}, where the system is trained specifically for predefined categories, and is not equipped to segment any others.}

Today there are two main strategies for generic object segmentation: saliency and object proposals.  
Both strategies capitalize on properties that can be learned from images and generalize to unseen objects (e.g., well-defined boundaries, differences with surroundings, shape cues, etc.).

\emph{Saliency methods} identify regions likely to capture human attention.  They yield either highly localized attention maps~\cite{LiuHZWL15,KruthiventiAB15,pan2016shallow,borji_survey} or a complete segmentation of the prominent object~\cite{zhang2013saliency, czmhh_contrastSaliency_cvpr11,PerazziKPH12,jiangsaliency,liu-salient,secret,zhao2015saliency,DeepSaliency}. Saliency focuses on regions that stand out, which is not the case for all foreground objects.

\begin{figure}[t]
\centering
\hspace*{-0.1in}
\begin{tabular}{c}
\includegraphics[width=1\columnwidth]{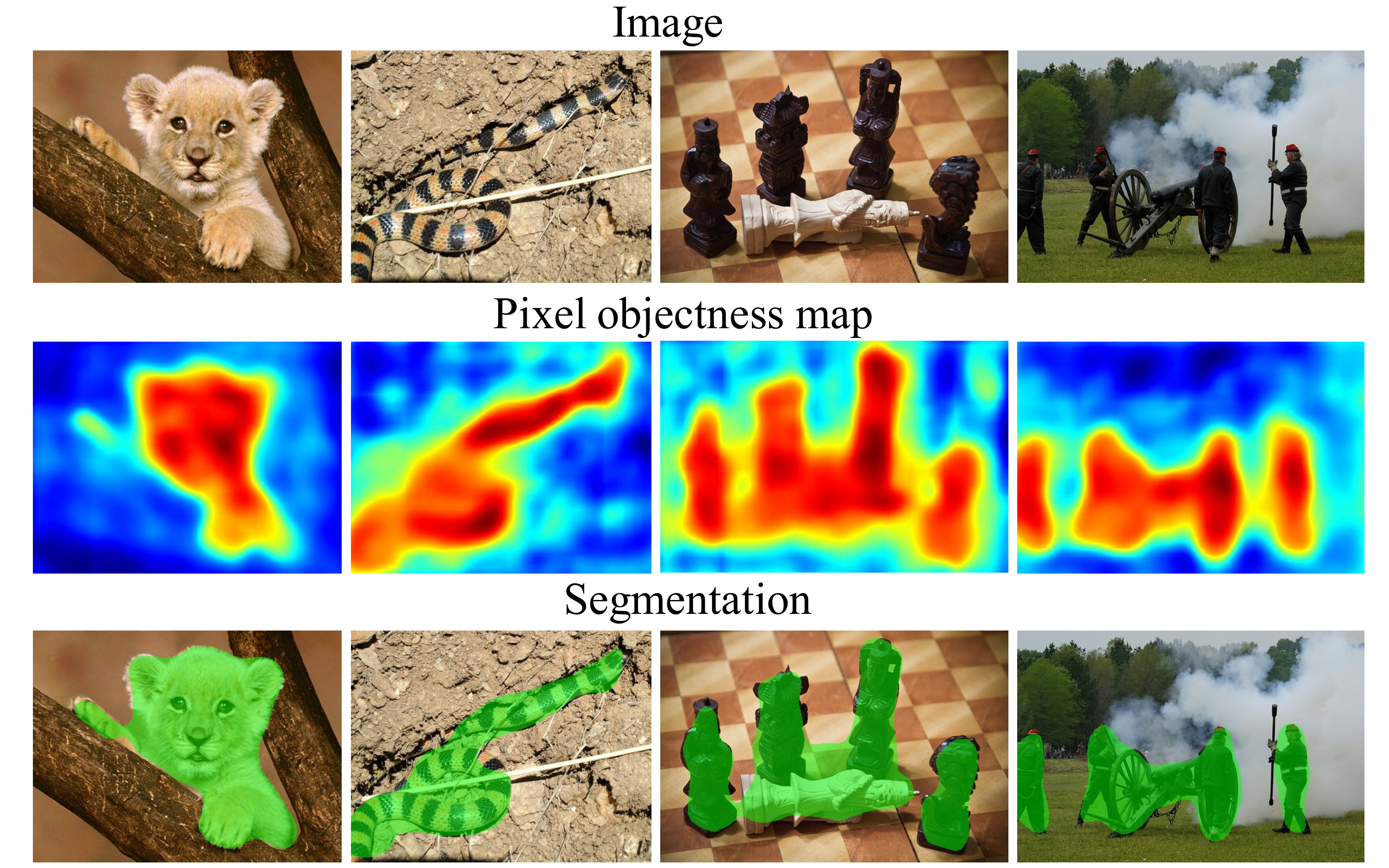}
\end{tabular}
  \captionsetup{ font={footnotesize}, skip=2pt}
\caption{\iccvf{Our method predicts an objectness map for each pixel (2nd row) and a single foreground segmentation (3rd row).  Left to right: It can accurately handle occluded objects, thin objects with similar colors to background, man-made objects, and even multiple objects.  It is class-independent and is not restricted to detect only particular objects.}}
\label{fig:concept}

\end{figure}

Alternatively, \emph{object proposal} methods learn to localize all objects in an image, regardless of their category~\cite{cpmc,APBMM2014,gop,endres,ZitnickECCV14edgeBoxes,UijlingsIJCV2013,deepmask,Hosang2015Pami}. The aim is to obtain high recall at the cost of low precision, i.e., they must generate a large number of proposals (typically 1000s) to cover all objects in an image. This usually involves a multi-stage process: first bottom-up segments are extracted, then they are scored by their degree of ``objectness".  Relying on bottom-up segments can be limiting, since low-level cues may fail to pull out contiguous regions for complex objects.  Furthermore, in practice, the accompanying scores  are not so reliable such that one can rely exclusively on the top \KG{few} proposals.

Motivated by these shortcomings, we introduce \emph{pixel objectness}, a new approach to generic foreground segmentation.  Given a novel image, the goal is to determine the likelihood that each pixel is part of a foreground object (as opposed to background or ``stuff" classes like grass, sky, sidewalks, etc.)  Our definition of a generic foreground object follows that commonly used in the object proposal literature~\cite{Alexe,cpmc,APBMM2014,gop,endres,ZitnickECCV14edgeBoxes,UijlingsIJCV2013}.  \iccvf{Pixel objectness generalizes window-level objectness~\cite{Alexe}. It quantifies how likely a pixel belongs to an object of \emph{any} class, and should be high even for objects unseen during training.}  See Fig.~\ref{fig:concept}.

We cast foreground object segmentation as a unified structured learning problem, and implement it by training a deep fully convolutional network to produce dense (binary) pixel label maps.  Given the goal to handle arbitrary objects, one might expect to need ample foreground-annotated examples across a vast array of categories to learn the generic cues.  However, we show that, somewhat surprisingly, when training with \emph{explicit boundary-level} annotations for few categories pooled together into a single generic ``object-like" class, pixel objectness generalizes well to \emph{thousands} of unseen objects.  This generalization ability is facilitated by an \emph{implicit image-level} notion of objectness built into a pretrained classification network, which we transfer to our segmentation model during initialization.

Our formulation has some key advantages.  First, it is not limited to segmenting objects that stand out conspicuously, as is often the case in salient object detection~\cite{czmhh_contrastSaliency_cvpr11,PerazziKPH12,jiangsaliency,liu-salient,secret,zhao2015saliency,DeepSaliency}.  Second, it is not restricted to segmenting only a fixed number of object categories, as is the case for supervised semantic segmentation~\cite{noh2015learning,crfasrnn_iccv2015,long_shelhamer_fcn,chen14semantic}. Third, unlike the two-stage processing typical in today's region proposal methods~\cite{cpmc,APBMM2014,gop,endres,ZitnickECCV14edgeBoxes,UijlingsIJCV2013}, our method unifies learning ``what makes a good region" with learning ``which pixels belong in a region together". Hence, it is not restricted to flawed regions from a bottom-up segmenter. 

\iccvf{Through extensive experiments, we show that our model generalizes very well to unseen objects.  We obtain state-of-the-art performance on the challenging ImageNet~\cite{imagenet_cvpr09} and MIT Object Discovery~\cite{rubinstein-cvpr2013} datasets. Finally, we show how to leverage our segmentations to benefit object-centric image retrieval and content-aware image resizing. In summary, we make the following novel contributions:}

\begin{itemize} 
	\item \iccvf{We are the first to show how to train a state-of-the-art generic object segmentation model without requiring a large number of annotated segmentations from thousands of diverse object categories.}
	\item \iccvf{Our novel formulation is neither restricted to a fixed set of categories (as in semantic segmentation) nor objects which stand out (as in saliency). It also unifies learning of grouping and objectness, unlike ``proposal" methods which treat them separately.}
	\item \iccvf{Through extensive results on 3,600$+$ categories and $\sim$1M images, our model generalizes to segment thousands of unseen categories. No other prior work---including recent deep saliency and object proposal methods---shows this level of generalization.}
\end{itemize}
\section{Related Work}

We divide related work into two top-level groups: (1) methods that extract an object mask no matter the object category, and (2) methods that learn from category-labeled data, and seek to recognize/segment those particular categories in new images.  Our method fits in the first group.

\subsection{Category-independent segmentation}

{\bf Interactive image segmentation} algorithms such as the popular GrabCut~\cite{grabcut} let a human guide the algorithm using bounding boxes or scribbles. These methods are most suitable when high precision segmentations are required such that some guidance from humans is worthwhile. While some methods try to minimize human involvement~\cite{icoseg,suyog-iccv2013}, still typically a human is always in the loop to guide the algorithm. In contrast, our model is fully automatic and segments foreground objects without any human guidance.

{\bf Object proposal methods}, also discussed above, produce thousands of generic object proposals either in the form of bounding boxes~\cite{endres,ZitnickECCV14edgeBoxes,UijlingsIJCV2013,ren2015faster} or regions~\cite{cpmc,APBMM2014,gop,deepmask,Hosang2015Pami}. Generating thousands of hypotheses ensures high recall, but often results in low precision.  Though effective for object detection, it is difficult to automatically filter out accurate proposals from this large hypothesis set without class-specific knowledge. We instead generate a \emph{single} hypothesis of the foreground as our final segmentation.  Our experiments directly evaluate our method's advantage.

{\bf Saliency models} have also been widely studied in the literature. The goal is to identify regions that are likely to capture human attention. While some methods produce highly localized regions~\cite{LiuHZWL15,KruthiventiAB15,pan2016shallow,borji_survey}, others segment complete objects~\cite{czmhh_contrastSaliency_cvpr11,PerazziKPH12,jiangsaliency,liu-salient,secret,zhao2015saliency,DeepSaliency}.  While saliency focuses on objects that ``stand out'',  our method is designed to segment all foreground objects, irrespective of whether they stand out in terms of low-level saliency. This is true even for the deep learning based saliency methods~\cite{pan2016shallow,KruthiventiAB15,LiuHZWL15,zhao2015saliency,DeepSaliency} which like us are end-to-end trained but prioritize objects that stand out.

\subsection{Category-specific segmentation}

{\bf Semantic segmentation} refers to the task of jointly \emph{recognizing} and segmenting objects, classifying each pixel into one of $k$ fixed categories. Recent advances in deep learning have fostered increased attention to this task. Most deep semantic segmentation models include fully convolutional networks that apply successive convolutions and pooling layers followed by upsampling or deconvolution operations in the end to produce pixel-wise segmentation maps~\cite{noh2015learning,crfasrnn_iccv2015,long_shelhamer_fcn,chen14semantic}. However, these methods are trained for a fixed number of categories. We are the first to  show that a fully convolutional network can be trained to accurately segment \emph{arbitrary} foreground objects.  Though \KG{relatively few} categories are seen in training, our model generalizes very well to unseen categories (as we demonstrate for 3,624 classes from ImageNet, only a fraction of which overlap with PASCAL, \KG{the source of our training masks}).

{\bf Weakly supervised joint segmentation} methods use weaker supervision than semantic segmentation methods. Given a batch of images known to contain the same object category, they segment the object in each one. The idea is to exploit the similarities within the collection to discover the common foreground. The output is either a pixel-level mask~\cite{classcut,vicente-cvpr2011,joulin-cvpr2012,kim-iccv2011,rubio-cvpr2012,rubinstein-cvpr2013,chen_cvpr14} or bounding box~\cite{ferrari_ijcv12,TangCVPR14}.  While joint segmentation is useful, its performance is limited by the shared structure within the collection; intra-class viewpoint and shape variations pose a significant challenge. Moreover, in most practical scenarios, such weak supervision is not available.  A \KG{stand alone} single-image segmentation model like ours is more widely applicable.

{\bf Propagation-based methods} transfer information from exemplars with human-labeled foreground masks~\cite{Ahmed_2014_CVPR,Yang_2015_CVPR,Kuettel2012cvpr,guillaumin2014imagenet,suyog-cvpr2016}. They usually involve a matching stage between likely foreground regions \KG{and} the exemplars. The downside is the need to store a large amount of exemplar data at test time and perform an expensive and potentially noisy matching process for each test image. In contrast, our segmentation model, once trained end-to-end, is very efficient to apply and does not need to retain any training data.

\section{Approach}
\label{sec:approach}

Our goal is to design a model that can predict the likelihood of each pixel being a generic foreground object as opposed to background. Building on the terminology from~\cite{Alexe}, we refer to our task as \emph{pixel objectness}. We use this name to distinguish our task from the related problems of salient object detection (which seeks only the most attention-grabbing foreground object) and region proposals (which seeks a ranked list of candidate object-like regions).  We pose pixel objectness as a dense labeling problem, and propose a solution based on a convolutional neural network architecture that supports end-to-end training.

First we introduce our core approach (Sec.~\ref{sec:dense_pred}). Then, we explore two applications that illustrate the utility of pixel objectness (Sec.~\ref{sec:apps}).

\subsection{Predicting Pixel Objectness}\label{sec:dense_pred}

{\bf Problem formulation:}  Given an RGB image$~\mathcal{I}$ of size $m \times n \times c$ as input, we formulate the task of foreground object segmentation as densely labeling each pixel in the image as either ``object" or ``background".  Thus the output of pixel objectness is a binary map of size $m \times n$.

Since our goal is to predict objectness for each pixel, our model should 1) predict a pixel-level map that aligns well with object boundaries, and 2) generalize so it can assign high probability to pixels of unseen object categories.

{\bf Challenges in dense foreground-labeled training data: }Potentially, one way to address both challenges would be to rely on a large annotated image dataset that contains a large number of diverse object categories with pixel-level foreground annotations.  However, such a dataset is non-trivial to obtain.  The practical issue is apparent looking at recent large-scale efforts to collect segmented images.  They contain boundary-level annotations for merely dozens of categories (20 in PASCAL~\cite{Everingham2010}, 80 in COCO~\cite{LinECCV14coco}), and/or for only a tiny fraction of all dataset images (0.03\% of ImageNet's 14M images have such masks).  Furthermore, such annotations come at a price---about \$400,000 to gather human-drawn outlines on 2.5M object instances from 80 categories~\cite{LinECCV14coco} assuming workers receive minimum wage.  To naively train a \emph{generic} foreground object segmentation system, one might expect to need foreground labels for many more representative categories, suggesting an alarming start-up annotation cost.

{\bf Mixing explicit and implicit representations of objectness:} This challenge motivates us to consider a different means of supervision to learn generic pixel objectness.  
Our idea is to train the system to predict pixel objectness using a mix of \emph{explicit} boundary-level annotations and \emph{implicit} image-level object category annotations.  From the former, the system will obtain direct information about image cues indicative of generic foreground object boundaries.   From the latter, it will learn object-like features across a wide spectrum of object types---but \emph{without} being told where those objects' boundaries are.

To this end, we propose to train a fully convolutional deep neural network for the foreground-background object labeling task.  We initialize the network using a powerful generic image representation learned from millions of images labeled by their object category, but lacking any foreground annotations.  Then, we fine-tune the network to produce dense binary segmentation maps, using \KG{relatively few images with pixel-level annotations originating from a small number of object categories.   }

Since the pretrained network is trained to recognize thousands of objects, we hypothesize that its image representation has a strong notion of objectness built inside it, even though it never observes \emph{any} segmentation annotations. 
Meanwhile, by subsequently training with explicit dense foreground labels, we can steer the method to fine-grained cues about boundaries that the standard object classification networks have no need to capture.   This way, even if our model is trained with a limited number of object categories having pixel-level annotations, we expect it to learn generic representations helpful to pixel objectness.

Specifically, we adopt a deep network structure~\cite{chen14semantic} originally designed for multi-class semantic segmentation.  We initialize it with weights pre-trained on ImageNet, which provides a representation equipped to perform image-level classification for some 1,000 object categories.  Next, we take a modestly sized semantic segmentation dataset, and transform its dense semantic masks into binary object vs.~background masks, by fusing together all its 20 categories into a single supercategory (``generic object").  We then train the deep network (initialized for ImageNet object classification) to perform well on the dense foreground pixel labeling task.  Our  model supports end-to-end training.

 \begin{figure}[t]
\centering
\renewcommand{\tabcolsep}{0pt}
  \captionsetup{ font={footnotesize}, skip=2pt}

\includegraphics[width=1\columnwidth]{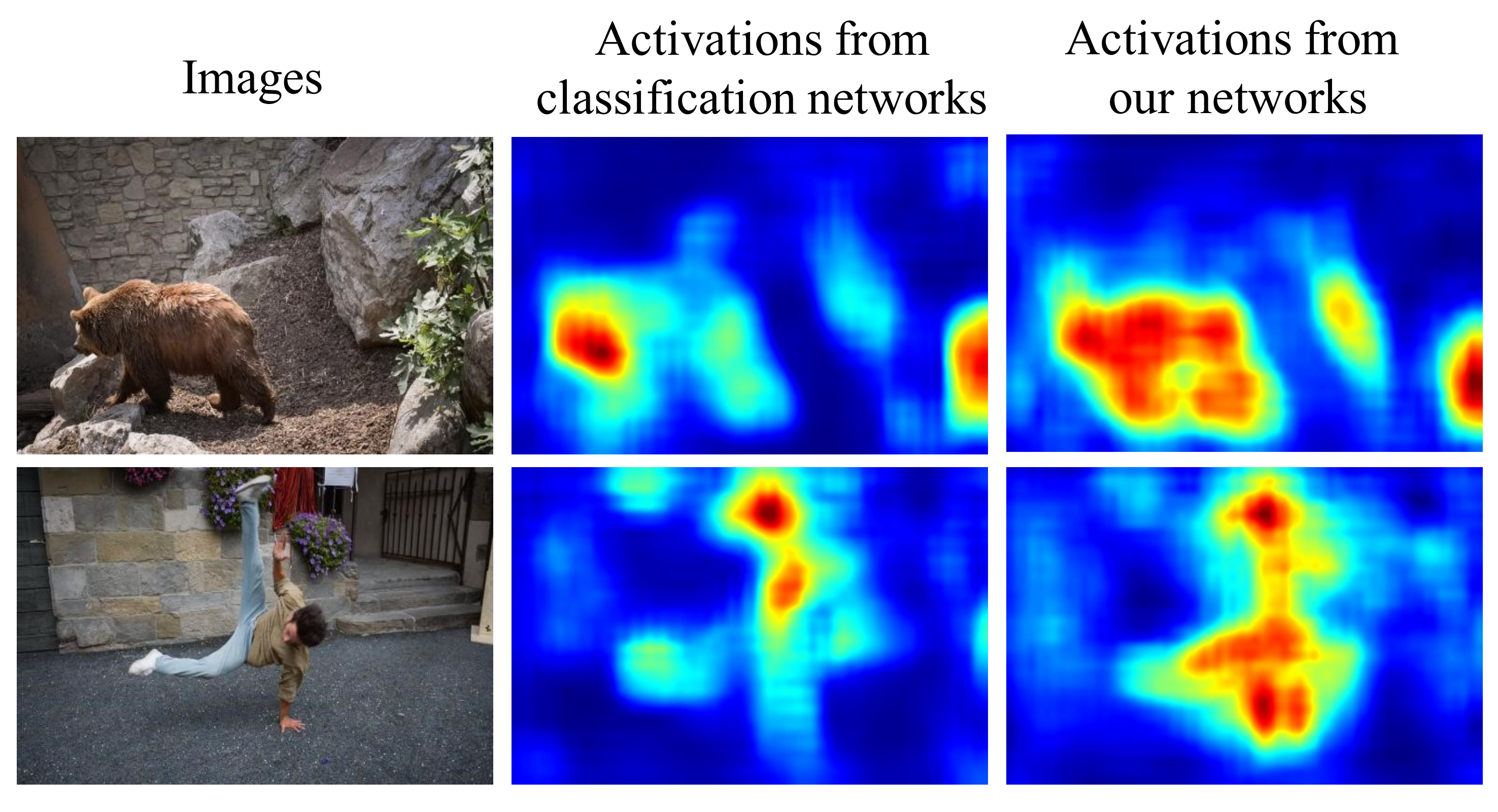}
\caption{Activation maps from a network (VGG~\cite{simonyan2014very}) trained for the classification task and our network which is fine-tuned with explicit dense foreground labels. We see that the classification network has already learned image representations that have \KG{some} notion of objectness, \KG{but with poor ``over''-localization}. Our network \KG{deepens the notion of objectness to pixels} and captures fine-grained cues about boundaries (best viewed on pdf).}
\label{fig:activation}
\end{figure}

\KG{To illustrate this synergy}, Fig.~\ref{fig:activation} shows activation maps from a network trained for ImageNet 
classification (middle) and from our network (right), by summing up feature 
responses from each filter in the last convolutional layer (pool5) for each spatial location. 
Although networks trained on a classification task never observe any segmentations, they can show high activation responses when object parts are present and low activation responses to 
stuff-like regions such as rocks and roads. Since the classification networks are trained with 
thousands of object categories, their activation responses \KG{are rather general}. However, they are responsive to \KG{only fragments of the objects}.   After training with explicit dense foreground labels, our network is able to extend high activation responses from 
discriminative object parts to the entire object.  

For example, in Fig.~\ref{fig:activation},  the classification 
network only has a high activation response on the bear's head, whereas  our pixel objectness network has a high response on the entire bear body; similarly for the person.  This \KG{supports} our hypothesis that networks trained for classification tasks \BX{contain} \KG{a reasonable but incomplete basis for objectness, despite lacking any spatial annotations}.  By subsequently training with explicit dense foreground labels, we can steer towards fine-grained cues about boundaries that the standard object classification networks have no need to capture.

{\bf Model architecture:} We adapt the widely used image classification model VGG-16 network~\cite{simonyan2014very} into a fully convolutional network by transforming its fully connected layers into convolutional layers~\cite{long_shelhamer_fcn,chen14semantic}. This enables the network to accept input images of any size and also produce corresponding dense output maps. The network comprises of stacks of convolution layers with max-pooling layers in between. All convolution filters are of size  $3 \times 3$ except the last convolution layer which comprises  $1 \times 1$ convolutions. Each convolution layer is also followed by a ``relu" non-linearity before being fed into the next layer. We remove the 1000-way classification layer from VGG-net and replace it with a 2-way layer that produces a binary mask as output.  The \KG{loss} is the sum of cross-entropy terms over each pixel in the output layer.  

The VGG-16 network consists of five max pooling layers.  While well suited for classification, this leads to a 32$\times$ reduction in the output resolution compared to the original image.  In order to achieve more fine-grained pixel objectness map, we apply the ``hole" algorithm proposed in~\cite{chen14semantic}. In particular, we replace the subsampling in the last two max-pooling layers with dilated convolutions~\cite{chen14semantic}.  This method is parameter free, results in only a 8$\times$ reduction in the output resolution and still retains a large field-of-view. We then use  bilinear interpolation to recover a foreground map at the original resolution. \KG{See appendix~for more details.} 

{\bf Training details:} To generate the explicit boundary-level training data, we rely on the 1,464 PASCAL 2012 segmentation training images~\cite{Everingham2010} \KG{and the additional annotations of~\cite{BharathICCV2011}, for 10,582 total training images.  The 20 object labels are discarded, and mapped instead to the single generic ``object-like'' (foreground) label for training.} We train our model using the Caffe implementation of~\cite{chen14semantic}. We optimize with stochastic gradient with a mini-batch size of 10 images. A simple data augmentation through mirroring the input images is also employed. A base learning rate of 0.001 with a $1/10$th slow-down every 2000 iterations is used. We train the network for a total of 10,000 iterations; total training time was about 8 hours on a modern GPU.

\subsection{Leveraging Pixel Objectness}\label{sec:apps}

Dense pixel objectness has many applications. \iccvf{Here we explore how it can assist in image retrieval and content-aware image retargeting, both of which demand a single, high-quality estimate of the foreground object region.}

\textbf{Object-aware image retrieval:} First, we consider how pixel objectness foregrounds can assist in image retrieval.  A retrieval system  accepts a query image containing an object, and then the system returns a ranked list of images that contain the same object. This is a valuable application, for example, to allow object-based online product search. Typically retrieval systems extract image features from the entire query image. This can be problematic, however, because it might retrieve images with similar background, especially when the object of interest is small.  We aim to use \KG{pixel objectness} to restrict the system's attention to the foreground object(s) as opposed to the entire image. 

To implement the idea, we first \KG{run pixel objectness}. In order to reduce false positive segmentations, we keep the largest connected foreground region if it is larger than $6\%$ of the overall image area. Then we crop the smallest bounding box enclosing the foreground segmentation and extract features from the entire bounding box. If no foreground is found (which occurs in roughly 17\% of all images), we extract image features from the entire image. The method is applied to both the query and database images. \KG{To rank database images,} we explore two image representations. The first one uses only the image features extracted from the bounding box, and the second concatenates the features from the original image with those from the bounding box.

\textbf{Foreground-aware image retargeting:} As a second application, we explore how \KG{pixel objectness} can enhance image retargeting. The goal is to adjust the aspect ratio or size of an image without distorting its important visual concepts. We build on the popular Seam Carving algorithm~\cite{avidan2007seam}, which eliminates the optimal irregularly shaped path, called a seam, from the image via dynamic programming.  In~\cite{avidan2007seam}, the energy is defined in terms of the image gradient magnitude. However, the gradient \KG{is not always} a sufficient energy function, especially when important visual content is non-textured or the background is textured.

Our idea is to protect semantically important visual content based on foreground segmentation.  
To this end, we consider a simple adaption of Seam Carving.  We define an energy function based on high-level semantics rather than low-level image features alone. Specifically, we first predict pixel objectness, and then we scale the gradient energy $g$ within the foreground segment(s) by $(g +1) \times 2$.

\section{Results}\label{sec:results}
We evaluate pixel objectness by comparing it to 16 recent methods in the literature, and also examine its utility for the two applications presented above. 

\noindent {\bf Datasets:} We \KG{use} three datasets \iccvf{which are commonly used to evaluate foreground object segmentation in images:}
\begin{itemize}
\item {\bf MIT Object Discovery:} This dataset consists of Airplanes, Cars, and Horses~\cite{rubinstein-cvpr2013}. It is most commonly used to evaluate weakly supervised segmentation methods. The images were primarily collected using internet search and the dataset comes with per-pixel ground truth segmentation masks. 
\item {\bf ImageNet-Localization:} We conduct a large-scale evaluation of our approach using ImageNet~\cite{ILSVRC15} ($\sim$1M images with bounding boxes, 3,624 classes). The diversity of this dataset lets us test the generalization abilities of our method. 
\item {\bf ImageNet-Segmentation:} This dataset contains 4,276 images from 445 ImageNet classes with \KG{pixel-wise} ground truth from~\cite{guillaumin2014imagenet}.

\end{itemize}

\noindent {\bf Baselines:} We compare to these state-of-the-art methods:
\begin{itemize}
\item {\bf Saliency Detection:} \KG{We compare to four salient object detection} methods~\cite{zhang2013saliency,jiangsaliency,zhao2015saliency,DeepSaliency}, selected for their efficiency and state-of-the-art performance. All these  methods are designed to produce a complete segmentation of the prominent object (vs.~fixation maps; see Sec. 5 of~\cite{zhang2013saliency}) and output continuous saliency maps, which are then thresholded by per image mean to obtain the segmentation.\footnote{This thresholding strategy was chosen because it gave the best results.}

\item {\bf Object Proposals:} We also compare with state-of-the-art region proposal algorithms, multiscale combinatorial grouping (MCG)~\cite{APBMM2014} and DeepMask~\cite{deepmask}. These methods output a ranked list of generic object segmentation proposals. The top ranked proposal in each image is taken as the final foreground segmentation for evaluation. We also compare with SalObj~\cite{secret} which uses saliency to merge multiple object proposals from MCG into a single foreground.

\item {\bf Weakly supervised joint-segmentation methods:} \KG{These approaches} rely on additional weak supervision in the form of prior knowledge that all images in a given collection share a common object category~\cite{rubinstein-cvpr2013,chen_cvpr14,joulin-cvpr2010,joulin-cvpr2012,kim-iccv2011,TangCVPR14,suyog-cvpr2016}. \KG{Note that our method lacks this additional supervision.}
\end{itemize}

\noindent {\bf Evaluation metrics:} Depending on the dataset, we use: (1) {\bf Jaccard Score:} Standard intersection-over-union (IoU) metric between predicted and ground truth segmentation masks and (2) {\bf BBox-CorLoc Score:} Percentage of objects correctly localized with a bounding box according to PASCAL criterion (i.e IoU $>$ 0.5) used in \cite{TangCVPR14,ferrari_ijcv12}.

For MIT and ImageNet-Segmentation, we use the segmentation masks and evaluate using the Jaccard score.  
For ImageNet-Localization we evaluate with the BBox-CorLoc metric, following the setup from~\cite{TangCVPR14,ferrari_ijcv12},
\cc{which entails putting a tight bounding box around our method's output.}

\subsection{Foreground object segmentation results}

\noindent {\bf MIT Object Discovery:} First we present results on the MIT dataset~\cite{rubinstein-cvpr2013}.   We do separate evaluation on the complete dataset and also a subset defined in~\cite{rubinstein-cvpr2013}. We compare our method with 13 existing state-of-the-art methods including saliency detection~\cite{zhang2013saliency,jiangsaliency,zhao2015saliency,DeepSaliency}, object proposal generation~\cite{APBMM2014,deepmask} plus merging~\cite{secret} and joint-segmentation~\cite{rubinstein-cvpr2013,chen_cvpr14,joulin-cvpr2010,joulin-cvpr2012,kim-iccv2011,suyog-cvpr2016}. We compare with author-reported results for the joint-segmentation baselines, and use software provided by the authors for the saliency and object proposal baselines.

\iccvf{Table~\ref{tab:results_mit} shows the results. Our proposed method outperforms several state-of-the-art saliency and object proposal methods---including recent deep learning techniques~\cite{zhao2015saliency,DeepSaliency,deepmask} in three out of six cases, and is competitive with the best performing method in the others.}

Our gains over the joint segmentation methods are arguably even more impressive because our model simply segments a single image at a time---no weak supervision!---and still substantially outperforms all weakly supervised techniques. \KG{We stress that} in addition to the weak supervision in the form of segmenting common object, the \KG{previous} best performing method~\cite{suyog-cvpr2016} also makes use of a pre-trained deep network; \KG{we use strictly less total supervision than~\cite{suyog-cvpr2016} yet still perform better.}  
Furthermore, most joint segmentation methods involve \KG{expensive} steps such as dense correspondences~\cite{rubinstein-cvpr2013} or region matching~\cite{suyog-cvpr2016} which can take up to hours \KG{even for a modest collection of 100 images.} In contrast, our method directly outputs the final segmentation in a single forward pass over the network and takes only 0.6 seconds \KG{ per image for complete processing.} \\ 

\begin{table}[t]
\centering
     	\captionsetup{width=0.48\textwidth, font={footnotesize}, skip=2pt}
     	   \scriptsize
     	   \setlength\tabcolsep{4pt}
   		\begin{tabular}{|c|c|c|c|c|c|c|}
   			\hline
     		\multirow{2}{*}{{\bf Methods}} & \multicolumn{3}{c|}{{\bf MIT dataset (subset)}} & \multicolumn{3}{c|}{{\bf MIT dataset (full)}} \\
\cline{2-7}
   			 & {\bf Airplane} & {\bf Car} & {\bf Horse} & {\bf Airplane} & {\bf Car} & {\bf Horse} \\
   			 \hline
   			 \hline
   							{\bf \# Images} & 82 & 89  & 93 & 470 & 1208 & 810 \\
   			\hline					
   \hline
	     \multicolumn{7}{|c|}{{\bf Joint Segmentation}} \\
   \hline
                            Joulin et al. \cite{joulin-cvpr2010} & 15.36 & 37.15  & 30.16 & n/a & n/a & n/a \\
\hline
   							Joulin et al. \cite{joulin-cvpr2012} & 11.72 & 35.15  & 29.53 & n/a & n/a & n/a \\
\hline
   							Kim et al. \cite{kim-iccv2011} & 7.9 &  0.04 & 6.43  & n/a & n/a & n/a \\
\hline
   							Rubinstein et al. \cite{rubinstein-cvpr2013} & 55.81 & 64.42 & 51.65  & 55.62 & 63.35  & 53.88 \\
\hline
   							Chen et al. \cite{chen_cvpr14} & 54.62 & 69.2  & 44.46   & 60.87 & 62.74 &  60.23 \\
\hline
   						Jain et al.~\cite{suyog-cvpr2016} & 58.65 & 66.47  & 53.57   & 62.27 &  65.3 & 55.41 \\
   \hline
   \hline
   	   
   	    \multicolumn{7}{|c|}{{\bf Saliency}} \\
   \hline
							    Jiang et al. \cite{jiangsaliency} & 37.22 & 55.22  & 47.02 & 41.52 & 54.34 & 49.67 \\
   \hline
      							Zhang et al. \cite{zhang2013saliency} & 51.84 & 46.61  & 39.52 & 54.09 & 47.38 & 44.12 \\
     \hline
						      DeepMC ~\cite{zhao2015saliency} & 41.75  & 59.16   & 39.34  & 42.84 &	58.13 &	41.85 \\  
     \hline
						      DeepSaliency~\cite{DeepSaliency} & 69.11 & 83.48  & 57.61 & {\bf 69.11} &	83.48	& {\bf 67.26} \\ 
     \hline
     \hline
        	    \multicolumn{7}{|c|}{{\bf Object Proposals}} \\
        \hline
				 MCG \cite{APBMM2014} & 32.02 & 54.21  & 37.85 & 35.32 & 52.98 & 40.44 \\
		\hline 
			    DeepMask~\cite{deepmask} & {\bf 71.81} & 67.01  & 58.80 &    68.89 &	65.4 &	62.61 \\
		\hline
				 SalObj \cite{secret} & 53.91 & 58.03  & 47.42 &  55.31 &	55.83	& 49.13 \\
\hline
\hline
  {\bf Ours} & 66.43 & {\bf 85.07}  & {\bf 60.85} & 66.18 & {\bf 84.80} & 64.90 \\
\hline
   				\end{tabular}
   				\caption{Quantitative results on MIT Object Discovery dataset. Our method outperforms several state-of-the-art methods for saliency detection, object proposals, and joint segmentation. (Metric: Jaccard score).}
  		\label{tab:results_mit}
\end{table}

\noindent {\bf ImageNet-Localization:} Next we present results on the ImageNet-Localization dataset. This involves testing our method on about 1 million images from 3,624 object categories. This also lets us test how generalizable our method is to unseen categories, i.e., those for which the method  sees no foreground examples during training.

Table~\ref{tab:results_imagenet} (left) shows the results. When doing the evaluation over all categories, we compare our method with five methods which report results on this dataset~\cite{Alexe,TangCVPR14,suyog-cvpr2016} or are scalable enough to be run at this large scale~\cite{jiangsaliency,APBMM2014}. We see that our method significantly improves the state-of-the-art.  The saliency and proposal methods~\cite{jiangsaliency,Alexe,APBMM2014} result in much poorer segmentations. Our method also significantly outperforms the joint segmentation approaches~\cite{TangCVPR14,suyog-cvpr2016}, which are the current best performing methods on this dataset. In terms of the actual number of images, our gains translate into correctly segmenting 42,900 more images than~\cite{suyog-cvpr2016} \KG{(which, like us, leverages ImageNet features)} and 83,800 more images than~\cite{TangCVPR14}. This reflects the overall magnitude of our gains over state-of-the-art baselines.

\begin{table}[t]
	\centering
	\scriptsize
	\setlength{\tabcolsep}{0.3em}
	\captionsetup{width=0.48\textwidth, font={footnotesize}, skip=2pt}
	\begin{tabular}{|c|c|c|}
		\hline
		\multicolumn{3}{|c|}{{\bf ImageNet-Localization dataset}} \\
		\hline
		\hline
		
		 & All & Non-Pascal \\
		\hline
		\hline
		\# Classes & 3,624 & 3,149 \\
		\hline
		\# Images & 939,516 & 810,219 \\
		\hline
		\hline
		
		Alexe et al. \cite{Alexe} & 37.42 & n/a \\
		\hline
		Tang et al. \cite{TangCVPR14}  &  53.20 & n/a \\
		\hline
		Jain et al.~\cite{suyog-cvpr2016}& 57.64 & n/a \\
		\hline
		\hline
		Jiang et al. \cite{jiangsaliency} & 41.28 & 39.35 \\
		\hline
		MCG \cite{APBMM2014} & 42.23 & 41.15 \\
		\hline
				\hline
				
		Ours & {\bf 62.12} & {\bf 60.18} \\
		\hline
	\end{tabular}
	\qquad
		\begin{tabular}{| c | c |}
			\hline
			\multicolumn{2}{|c|}{{\bf ImageNet-Segmentation dataset}} \\
			\hline		
			\hline
			
			Jiang et al. \cite{jiangsaliency} & 43.16  \\
			\hline	
			Zhang et al. \cite{zhang2013saliency} & 45.07  \\
			\hline	
			DeepMC ~\cite{zhao2015saliency} & 40.23  \\
			\hline	
			DeepSaliency~\cite{DeepSaliency} &   61.12 \\
			\hline	
			\hline
			MCG \cite{APBMM2014} & 39.97  \\
			\hline		
			DeepMask~\cite{deepmask} & 58.69   \\
			\hline
			SalObj~\cite{secret} & 41.35   \\
			\hline 
			Guillaumin et al.~\cite{guillaumin2014imagenet} & 57.3 \\
			\hline
			\hline
			Ours & {\bf 64.22} \\
			\hline			
		\end{tabular}
	
			\caption{Quantitative results on ImageNet localization and segmentation datasets. Results on ImageNet-Localization (left) show that the proposed model outperforms several state-of-the-art methods and also generalizes very well to unseen object categories (Metric: BBox-CorLoc). It also outperforms all methods on the ImageNet-Segmentation dataset (right) showing that it produces high-quality object boundaries (Metric: Jaccard score).}
	\label{tab:results_imagenet}
\end{table}

\begin{figure*}[t]
	\centering
	\renewcommand{\tabcolsep}{0pt}
	\captionsetup{width=1\textwidth, font={footnotesize}, skip=2pt}
	\begin{tabular}{c}
		ImageNet Examples from Pascal Categories \\
		\includegraphics[keepaspectratio=true,scale=0.37]{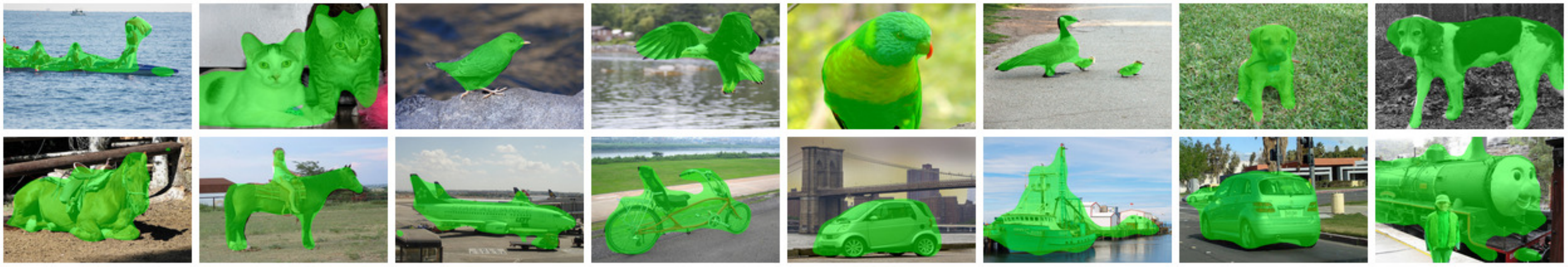} \\
		\hline
		ImageNet Examples from Non-Pascal Categories (unseen) \\
		\includegraphics[keepaspectratio=true,scale=0.37]{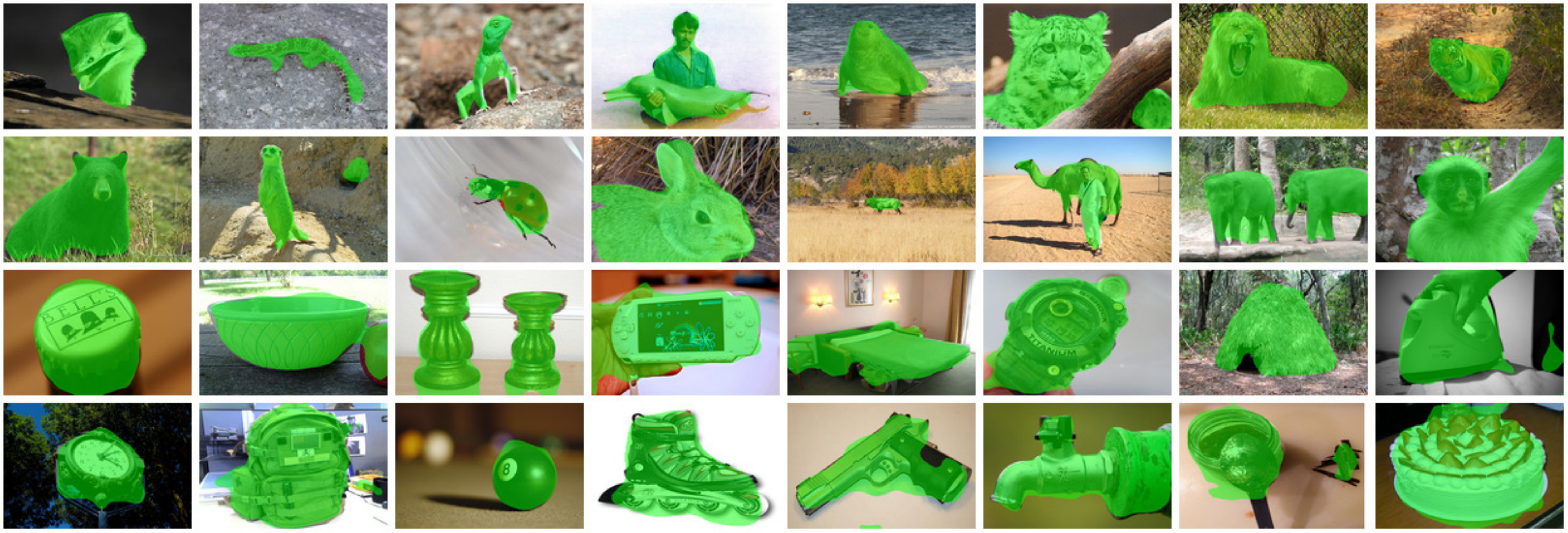} \\
		\hline
		Failure cases \\
		\includegraphics[keepaspectratio=true,scale=0.37]{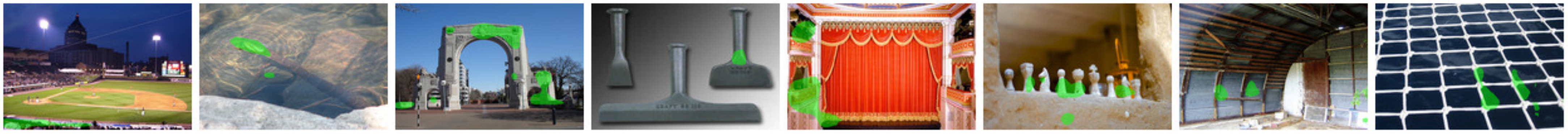} \\
	\end{tabular}
	\caption{Qualitative results: We show qualitative results on images belonging to PASCAL (top) and Non-PASCAL (middle) categories. Our segmentation model generalizes remarkably well even to those categories which were unseen \KG{in any foreground mask} during training (middle rows). Typical failure cases (bottom) involve scene-centric images where it is not easy to clearly identify foreground objects (best viewed on pdf). }
	\label{fig:qual_res}
	
\end{figure*}

Does our learned segmentation model  only recognize foreground objects that it has seen during training, or can it generalize to unseen object categories? Intuitively, ImageNet has such a large number of diverse categories that this gain would not have been possible if our method was only over-fitting to the 20 seen PASCAL categories.  To empirically verify this intuition, we next exclude those ImageNet categories which are directly related to the PASCAL objects, by matching the two datasets' synsets.  This results in a total of 3,149 categories which are exclusive to ImageNet (``Non-PASCAL'').   See Table~\ref{tab:results_imagenet} (left) for the data statistics.

We see only a very marginal drop in performance; our method still significantly outperforms both the saliency and object proposal baselines. This is an \KG{important} result, because during training the segmentation model \emph{never saw any dense object masks for images in these categories}. \KG{Bootstrapping from} the pretrained weights of the VGG-classification network, our model is able to learn a transformation between its prior belief on what looks like an object to complete dense foreground segmentations. \\

\noindent {\bf ImageNet-Segmentation: }
Finally, we measure the \KG{pixel-wise} segmentation quality on a large scale.  For this \KG{we use the} ground truth masks provided by~\cite{guillaumin2014imagenet} for 4,276 images from 445 ImageNet categories. \iccvf{The current best reported results are from the segmentation propagation approach of~\cite{guillaumin2014imagenet}. We found that DeepSaliency~\cite{DeepSaliency} and DeepMask~\cite{deepmask} further improve it. Note that like us, DeepSaliency~\cite{DeepSaliency} also trains with PASCAL~\cite{Everingham2010}. DeepMask~\cite{deepmask} is trained with a much larger COCO~\cite{LinECCV14coco} dataset. Our method outperforms all methods, significantly improving the state-of-the-art (see Table~\ref{tab:results_imagenet} (right)). This shows that our method not only generalizes to thousands of object categories but also produces high quality object segmentations.}

\begin{figure}[t]
	\centering
	\renewcommand{\tabcolsep}{0pt}
	\captionsetup{width=0.48\textwidth, font={footnotesize}, skip=2pt}
	\begin{tabular}{c}
		Images \\
		\includegraphics[width=0.95\columnwidth]{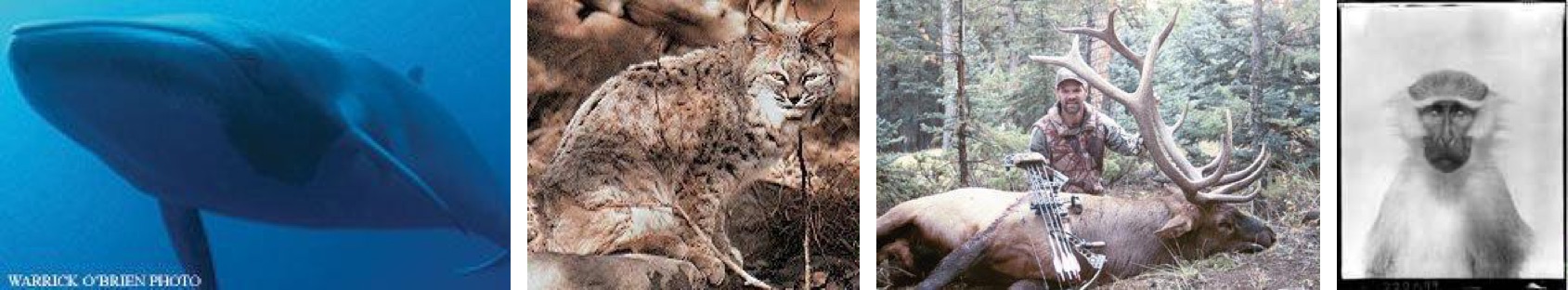} \\
		DeepSaliency~\cite{DeepSaliency} \\
		\includegraphics[width=0.95\columnwidth]{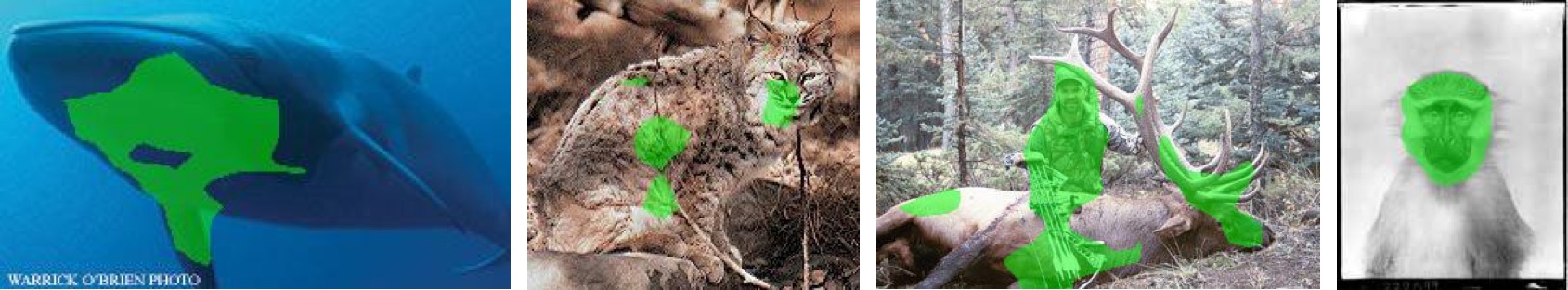} \\
		Ours \\
		\includegraphics[width=0.95\columnwidth]{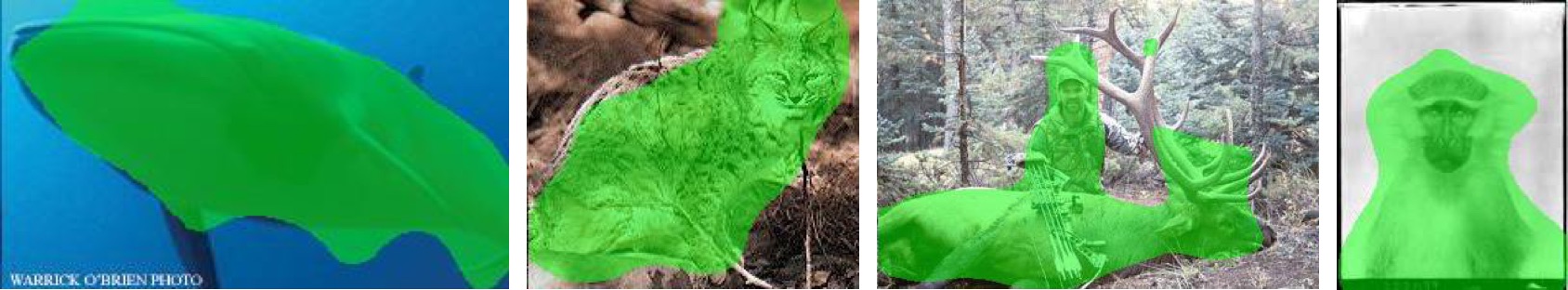} \\
		
	\end{tabular}
	\caption{Visual comparison for our method and the best performing saliency method, DeepSaliency~\cite{DeepSaliency}, which can fail when an object does not ``stand out" from background \KG{(best viewed on pdf). }}
	\label{fig:qual_sal}
	
\end{figure}

\noindent {\bf Pixel objectness vs.~saliency:} Salient object segmentation methods can potentially fail in cases where the foreground object does not stand out from the background. On the other hand, pixel objectness is designed to find objects even if they are not salient. To verify this hypothesis, we ranked all the images in the ImageNet-Segmentation dataset~\cite{guillaumin2014imagenet} by the appearance overlap between the foreground object and background. For this, we make use of the ground-truth segmentation to compute a 30-bin RGB color histogram for foreground and background respectively. We then compute cosine distance between the normalized histograms to measure how similar their distributions are and use that as a measure of separability between foreground and background.


\begin{figure}[t!]
	\centering
	\captionsetup{width=0.48\textwidth, font={footnotesize}, skip=2pt}
	\includegraphics[width=\columnwidth]{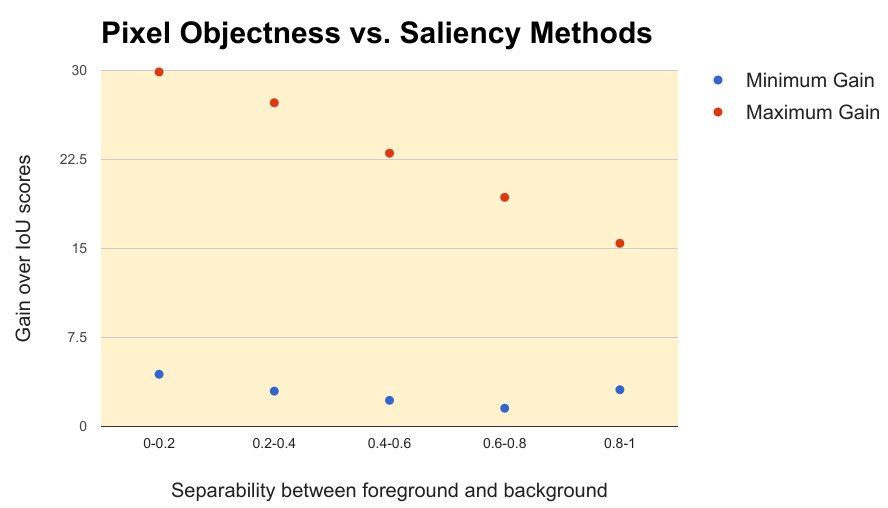}
	\caption{Pixel objectness vs. saliency methods: performance gains grouped using foreground-background separability scores. On the x-axis, lower scores mean that the objects are less salient and thus difficult to separate from background. On the y-axis, we plot the maximum and minimum gains of pixel objectness with other saliency methods. }
	\label{fig:saliency_th}
\end{figure}

Figure~\ref{fig:saliency_th} groups different images based on their separability scores and shows the minimum and maximum gains of our method over four state-of-the-art saliency methods~\cite{zhang2013saliency,jiangsaliency,zhao2015saliency,DeepSaliency} for each group. Lower separability score means that the foreground and background strongly overlaps and hence objects are not salient. First, we see that our method has positive gains across all groups showing that it outperforms all other saliency methods in every case. Secondly, we see that our gains are higher for lower separability scores. This demonstrates that the saliency methods are much weaker when foreground and background are not easily separable. On the other hand, pixel objectness works well irrespective of whether the foreground objects are salient or not. Our average gain over DeepSaliency~\cite{DeepSaliency} is 4.4\% IoU score on the subset obtained by thresholding at 0.2 (1320 images) as opposed to 3.1\% IoU score over the entire dataset.

Fig.~\ref{fig:qual_sal} visually illustrates this. Even the best performing saliency method~\cite{DeepSaliency} fails in cases where an object does not stand out from the background. In contrast, pixel objectness successfully finds complete objects even in these images.  

\vspace{5pt}
\noindent {\bf Qualitative results:} Fig.~\ref{fig:qual_res} shows qualitative results for ImageNet from both PASCAL and Non-PASCAL categories. \KG{Pixel objectness} accurately segments foreground objects from both sets. The examples from the Non-PASCAL categories highlight its strong generalization capabilities. \iccvf{We are able to segment objects across scales and appearance variations, including multiple objects in an image. It can segment even man-made objects, which are especially distinct from the objects in PASCAL (see appendix~for more examples). The bottom row shows failure cases. Our model has more difficulty in segmenting scene-centric images where it is more difficult to clearly identify foreground objects.}

\subsection{Impact on downstream applications}

Next we report results leveraging pixel objectness for two downstream tasks.
\subsubsection{Object-aware image retrieval} \label{sec:results_search}
First we consider the object-based image retrieval task \KG{defined in Sec.~\ref{sec:apps}}. We use the ILSVRC2012~\cite{ILSVRC15} validation set, which contains 50K images and $1,000$ object classes, with $50$ images per class. As an evaluation metric, we use mean average precision (mAP). We extract VGGNet~\cite{simonyan2014very} features and use cosine distance to rank retrieved images.

We compare with two baselines 1) \textbf{Full image}, which ranks images based on features extracted from the entire image, and 2) \textbf{Top proposal} (TP), which ranks images based on features extracted from the top ranked MCG~\cite{APBMM2014} proposal.  For our method and the Top proposal baseline, we examine two image representations. The first directly uses the features extracted from the region containing the foreground or the top proposal (denoted \textbf{FG}). The second representation concatenates the extracted features with the image features extracted from the entire image \KG{(denoted \textbf{FF})}.

\begin{figure}[t]
	\centering
	\captionsetup{width=0.48\textwidth, font={footnotesize}, skip=2pt}
	\includegraphics[width=1\columnwidth]{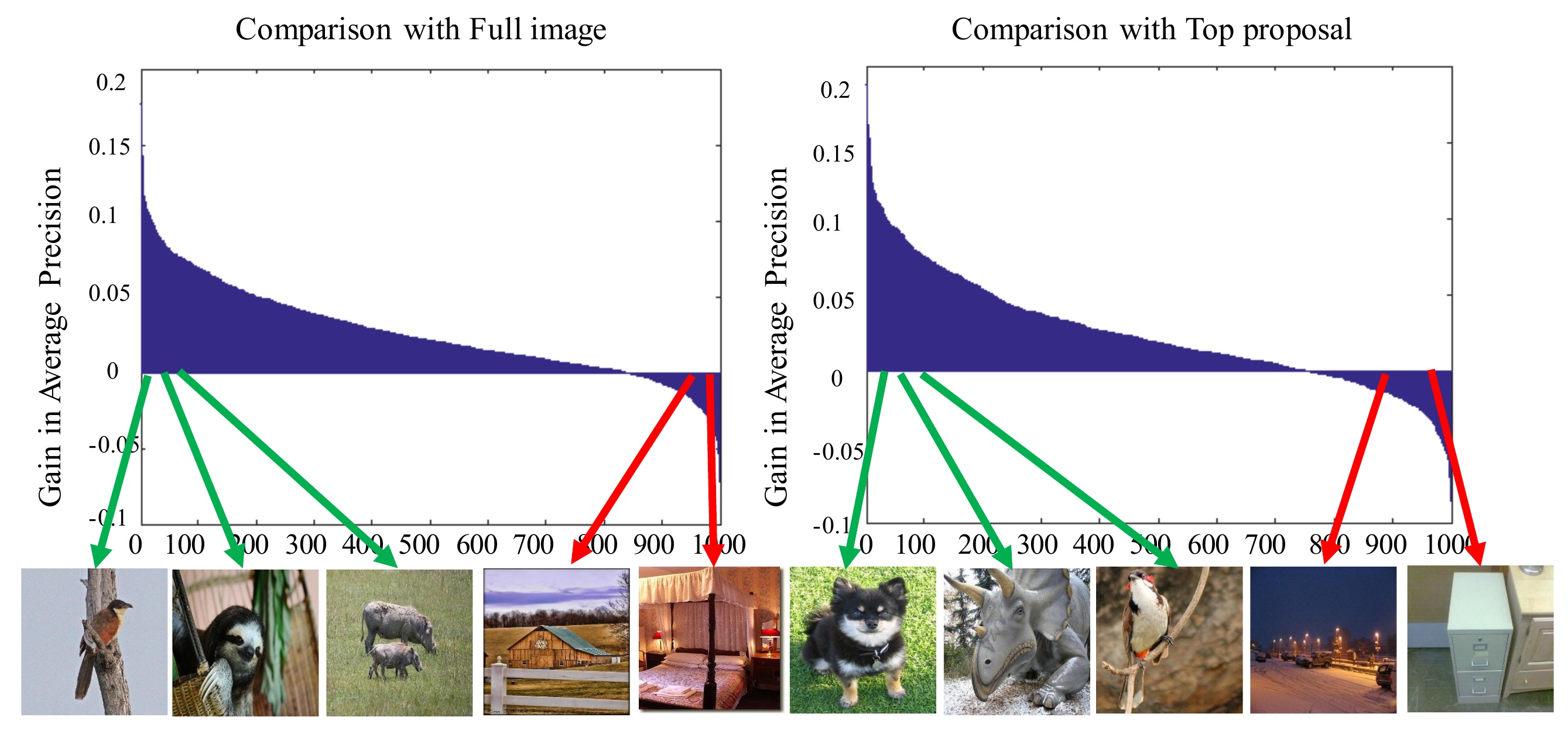}
	\caption{
		We show the \emph{gain} in average precision per object class between our method and the baselines (Full image on the left, and Top proposal on the right). Green arrows indicate example object classes for which our method performs better and red arrows indicate object classes for which the baselines perform better. Note our method excels at retrieving natural objects but can fail for scene-centric classes.}
	\label{fig:category}
	\vspace{5pt}
\end{figure}

Table~\ref{tab:result} shows the results. Our method with FF yields the best results. Our method outperforms both baselines for \KG{many} ImageNet classes. Figure~\ref{fig:category} looks more closely at the distribution of our method's gains in average precision per class.
We observe that our method performs extremely well on object-centric classes such as animals, but has limited improvement upon the baseline on scene-centric classes (lakeshore, seashore etc.).  To verify our hypothesis, we isolate the results on the first 400 object classes of ImageNet, which contain mostly \KG{ object-centric classes, as opposed to scene-centric objects.} On those first 400 object classes, our method outperforms both baselines by a larger margin (see appendix).  This demonstrates the value of our method at retrieving objects, which often contain diverse background and so naturally benefit more from accurate pixel objectness.

To further understand the superior performance of our method, we show the Top-5 nearest neighbors for both our method and the Full image baseline in Figure~\ref{fig:nn}. In the first example (first and second rows), the query image contains a small bird. Our method is able to segment the bird and retrieves relevant images that also contain birds. The baseline, on the contrary, \KG{has noisier retrievals due to mixing the background}. The last two rows show a case where, \KG{at least according to ImageNet labels,} our method fails.   Our method segments the person, and then retrieves images containing a person from different scenes, whereas the baseline focuses on the entire image and retrieves similar scenes.

\begin{table}[t]

\captionsetup{width=0.48\textwidth, font={footnotesize}, skip=2pt}
           \scriptsize
           \setlength\tabcolsep{4pt}
\centering
{\scriptsize
\hspace*{-0.1in}
\begin{tabular}{ |c|c|c|c|c|c|}
 \hline

Method& Ours(FF) &Ours(FG)& Full Img  & TP (FF)~\cite{APBMM2014} & TP (FG)~\cite{APBMM2014}  \\ \hline
All & \bf{0.3342}&0.3173 &  0.3082&0.3102&0.2092 \\ \hline
\BX{Obj-centric}  &\bf{0.4166}&0.4106  &0.3695&0.3734 &0.2679 \\ \hline

\end{tabular}}
\caption{Object-based image retrieval performance on ImageNet. We report average precision on the entire validation set, and on the first 400 categories, which are \BX{mostly object-centric classes}. }

\label{tab:result}
\end{table}

\begin{figure}[t]
\centering

\renewcommand{\tabcolsep}{0pt}
  \captionsetup{ font={footnotesize}, skip=2pt}

\includegraphics[width=0.9\columnwidth]{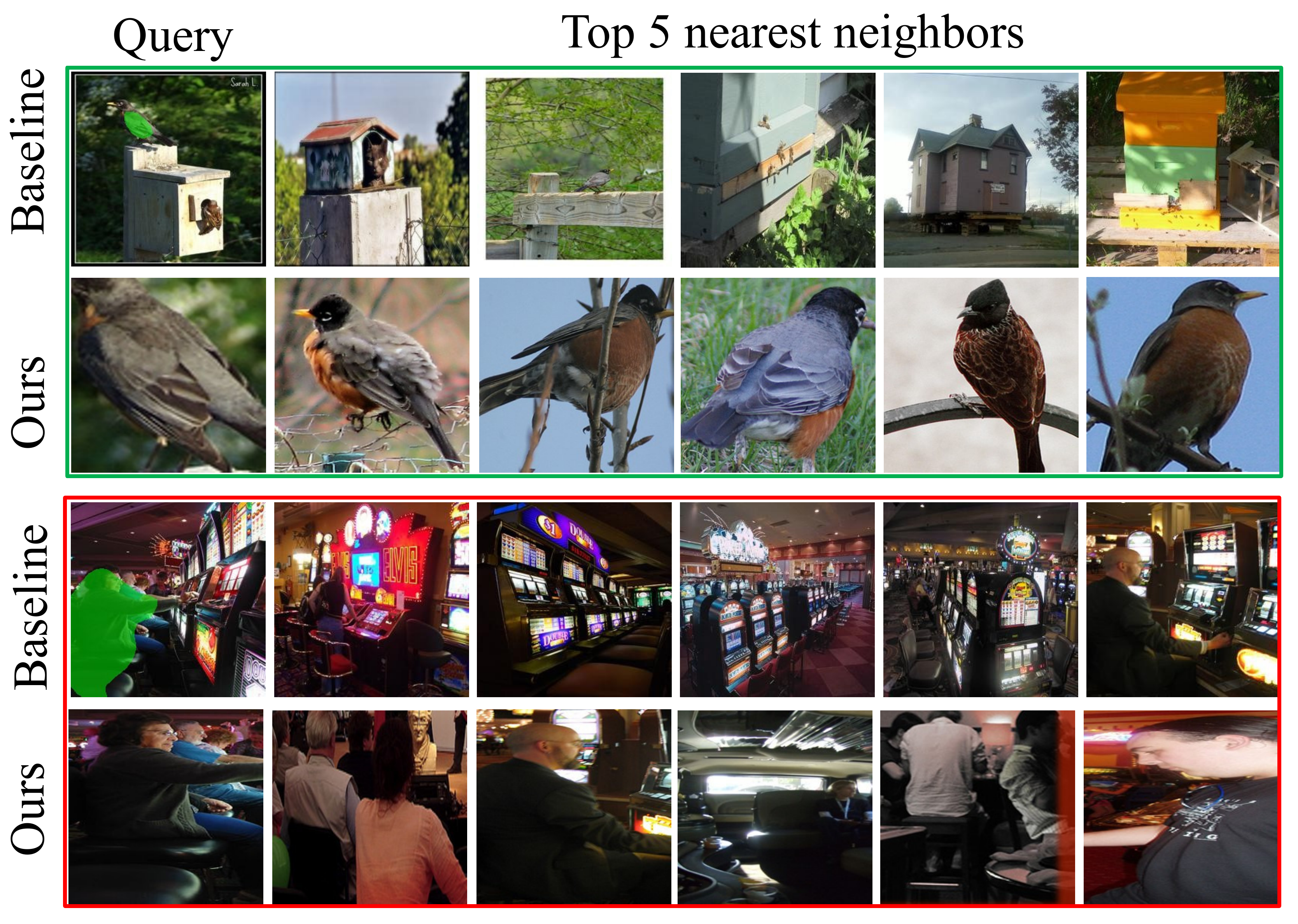}
\caption{Leveraging pixel objectness for object aware image retrieval (best viewed on pdf).}
\label{fig:nn}
\end{figure}

\subsubsection{Foreground-aware image retargeting} \label{sec:results_mturk}

Next, we show how to enhance Seam Carving retargeting with pixel objectness predictions. We use a random subset of 500 images from the 2014 Microsoft COCO Captioning Challenge Testing Images~\cite{LinECCV14coco} for experiments.

\begin{figure}[t]
  \centering
  \renewcommand{\tabcolsep}{0pt}
  \captionsetup{font={footnotesize}, skip=2pt}

    \includegraphics[width=0.98\columnwidth]{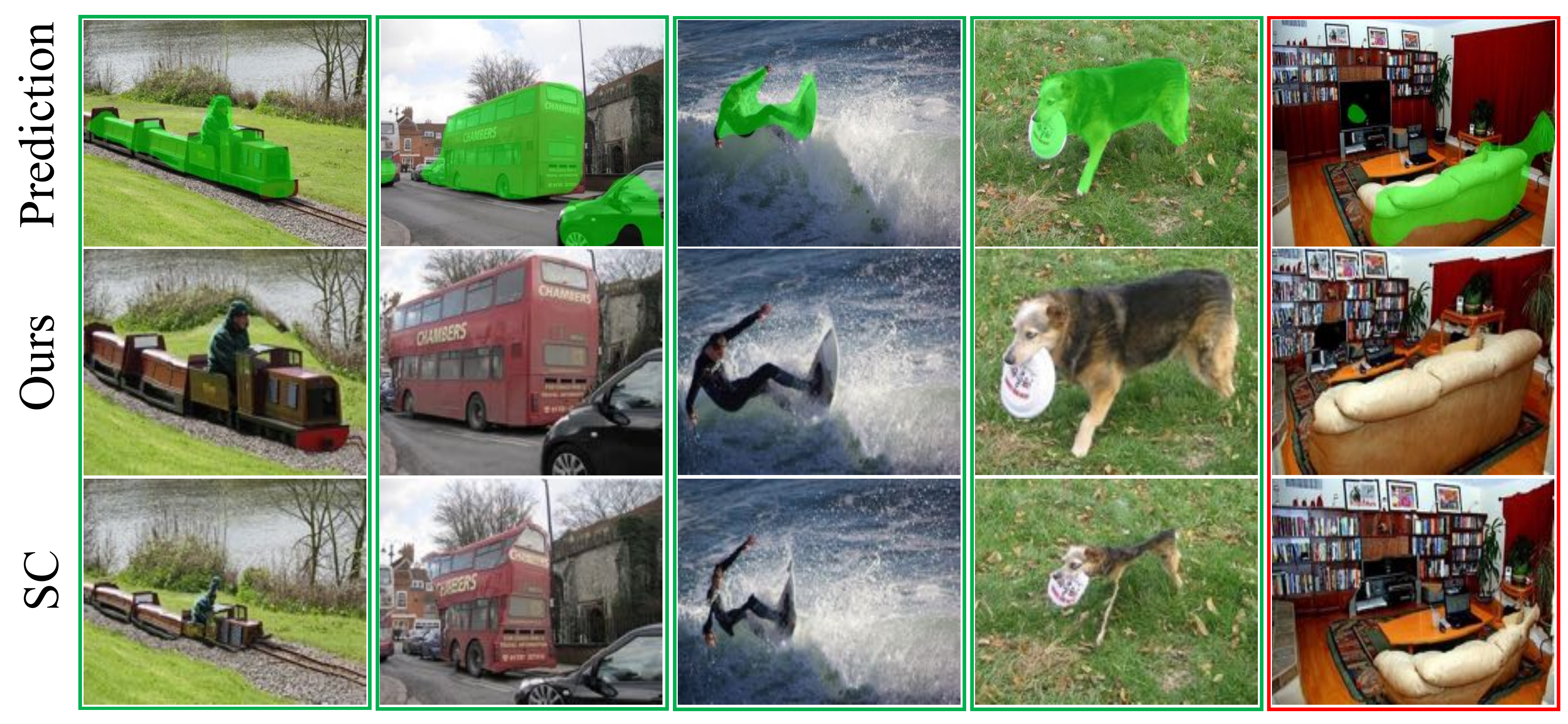} \\
   \caption{Leveraging pixel objectness for foreground aware image retargeting. See appendix for more examples  (best viewed on pdf).}
\label{fig:seam_example}
\end{figure}

Figure~\ref{fig:seam_example} shows example results. For reference, we also compare with the original Seam Carving (SC) algorithm \cite{avidan2007seam} that uses image gradients as  the energy function. Both methods are instructed to resize the source image to $2/3$ of its original size. Thanks to the proposed foreground segmentation, our method successfully preserves the important visual content (e.g., train, bus, human and dog) while reducing the content of the background. The baseline produces images with important objects distorted, because gradient strength is an inadequate indicator for perceived content, especially when background is textured. The rightmost column is a failure case for our method on a scene-centric image that does not contain any salient objects.

To quantify the results over all 500 images, we perform a human study on Amazon Mechanical Turk. We present image pairs produced by our method and the baseline in arbitrary order and ask workers to rank which image is more likely to have been manipulated by a computer.  Each image pair is evaluated by three different workers. Workers found that $38.53\%$ of the time images produced by our method are more likely to have been manipulated by a computer, $48.87\%$ for the baseline; both methods tie $12.60\%$ of the time. Thus, human evaluation with non-experts demonstrates that our method outperforms the baseline. In addition, we also ask a vision expert familiar with image retargeting---but not involved in this project---to score the 500 image pairs 
with the same interface as the crowd workers.  The vision expert found our method performs better for $78\%$ of the images, baseline is better for $13\%$, and both methods tie for $9\%$ images. This further confirms that our foreground prediction can enhance image retargeting by defining a more semantically meaningful energy function.

\section{Conclusions}

We proposed an end-to-end learning framework for segmenting generic foreground objects in images. Our results demonstrate its effectiveness, with significant improvements over the state-of-the-art on multiple datasets. Our results also show that \KG{pixel objectness} generalizes very well to thousands of unseen object categories. The foreground segmentations produced by our model also proved to be highly effective in improving the performance of image-retrieval and image-retargeting tasks, which helps illustrate the real-world demand for high-quality, single image, non-interactive foreground segmentations. \\

\noindent {\bf Code and pre-trained models available at:} \\ \url{http://vision.cs.utexas.edu/projects/pixelobjectness/} \\

\noindent {\bf Acknowledgements:} This research is supported in part by ONR YIP N00014-12-1-0754.

\bibliographystyle{IEEEtran}
\bibliography{generic_object_extraction}
\section{Appendix}

\subsection{CNN Architecture (Sec.~\ref{sec:dense_pred} in the main text)} Here we provide more details of the fully convolutional architecture that was employed to train our model for pixel objectness.
\vspace{5pt} \\
\noindent {\bf Notations: }

\begin{enumerate}
\item {\bf Convolution layers:} conv x-y denotes a convolution layer with x $\times$ x kernels and y channels, a stride of 1 was used everywhere.
\item {\bf Max Pooling:} maxpool denotes a max-pooling layer with KS as kernel size and stride S.
\item {\bf Non Linearity:} A relu non-linear activation function was used after each convolution layer.
\item {\bf Dropout:} Dropout regularization was used in the last layers with a ratio of 0.5.
\end{enumerate}

\noindent {\bf Architecture: }

\begin{itemize}
	\item Input Image: 3-channel RGB (3 $\times$ 321 $\times$ 321)
	\item conv3-64 $\ $ $\rightarrow$ relu $\rightarrow$ conv3-64 $\ $ $\rightarrow$ relu $\rightarrow$ maxpool (KS:3, S:2)
	\item conv3-128 $\rightarrow$ relu $\rightarrow$ conv3-128 $\rightarrow$ relu $\rightarrow$ maxpool (KS:3, S:2)
	\item conv3-256 $\rightarrow$ relu $\rightarrow$ conv3-256 $\rightarrow$ relu $\rightarrow$ conv3-256 $\rightarrow$ relu $\rightarrow$ maxpool (KS:3, S:2)
	\item conv3-512 $\rightarrow$ relu $\rightarrow$ conv3-512 $\rightarrow$ relu $\rightarrow$ conv3-512 $\rightarrow$ relu $\rightarrow$ maxpool (KS:3, S:1)
	\item conv3-512 $\rightarrow$ relu $\rightarrow$ conv3-512 $\rightarrow$ relu $\rightarrow$ conv3-512 $\rightarrow$ relu $\rightarrow$ maxpool (KS:3, S:1)
	\item conv3-1024 $\rightarrow$ relu $\rightarrow$ dropout (0.5)  $\rightarrow$ conv1-1024 $\rightarrow$ relu $\rightarrow$ dropout (0.5) $\rightarrow$ conv1-2
\end{itemize}

\noindent Even though this architecture largely follows the standard VGG-16~\cite{simonyan2014very} architecture, there are minor changes similar to~\cite{chen14semantic} which enables us to obtain a higher resolution output map. This modified network was initialized using the VGG-16 pre-trained weights provided by~\cite{chen14semantic}.

\subsection{More Qualitative Results of Pixel Objectness (Sec.~\ref{sec:results} in the main text)}
\begin{figure*}[h!]
  \centering
  \renewcommand{\tabcolsep}{0pt}
  \captionsetup{width=1\textwidth, font={footnotesize}, skip=2pt}
   \begin{tabular}{|c|}
      \hline
      {\bf Additional ImageNet Qualitative Examples from PASCAL Categories} \\
      \hline
    \includegraphics[width=1.8\columnwidth]{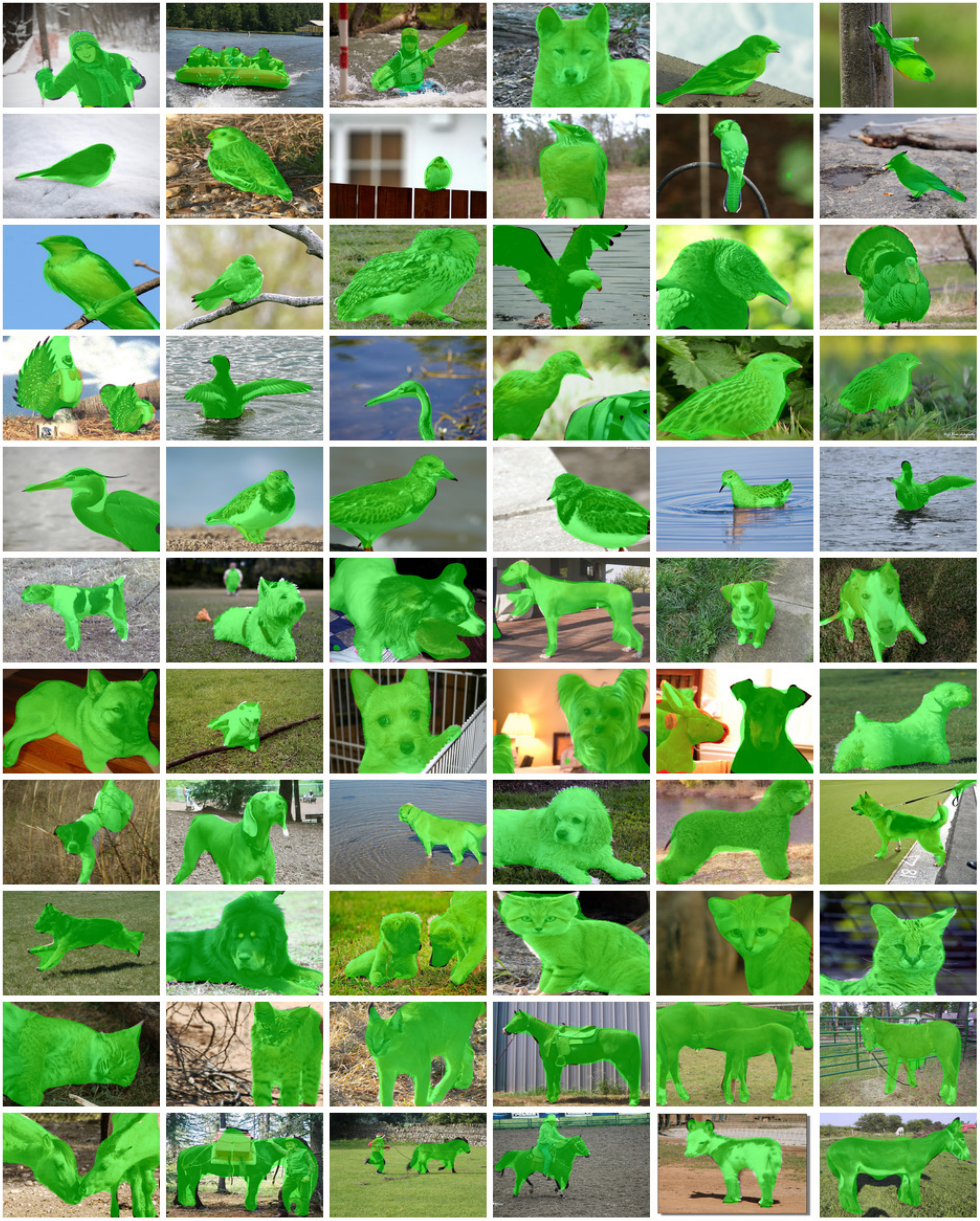} \\
	\hline
	\end{tabular}
		\caption{Qualitative results: We show example segmentations from ImageNet dataset obtained by our pixel objectness model. The segmentation results are shown with a green overlay. Our method is able to accurately segment foreground objects including cases where the objects are not highly salient. Best viewed in color.}
	  \label{fig:qual_res_seen1}
\end{figure*}

\begin{figure*}[h!]
  \centering
  \renewcommand{\tabcolsep}{0pt}
  \captionsetup{width=1\textwidth, font={footnotesize}, skip=2pt}
   \begin{tabular}{|c|}
      \hline
      {\bf Additional ImageNet Qualitative Examples from PASCAL Categories} \\
      \hline
    \includegraphics[width=1.8\columnwidth]{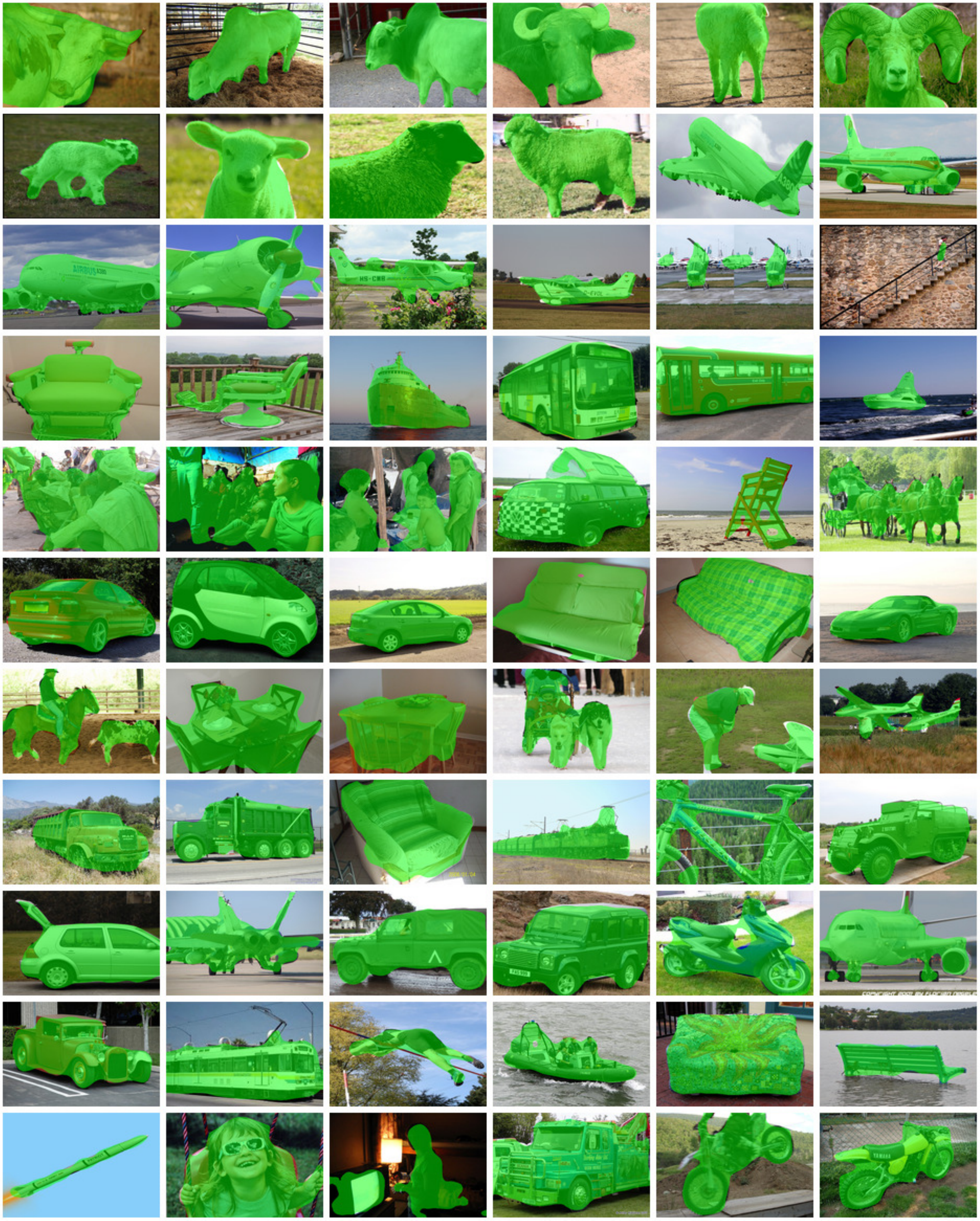} \\
	\hline
	\end{tabular}
	\caption{Qualitative results: We show example segmentations from ImageNet dataset obtained by our pixel objectness model on PASCAL Categories. The segmentation results are shown with a green overlay. Our method is able to accurately segment foreground objects including cases where the objects are not highly salient. Best viewed in color.}
	  \label{fig:qual_res_seen2}
\end{figure*}

We also show additional qualitative results from ImageNet dataset for our proposed pixel objectness model. Figure~\ref{fig:qual_res_seen1},~\ref{fig:qual_res_seen2} show the qualitative results for ImageNet images which belong to PASCAL categories. Our method is able to accurately segment foreground objects including cases with multiple foreground objects as well as the ones where the foreground objects are not highly salient.

\begin{figure*}[h!]
  \centering
  \renewcommand{\tabcolsep}{0pt}
  \captionsetup{width=1\textwidth, font={footnotesize}, skip=2pt}
   \begin{tabular}{|c|}
      \hline
      {\bf Additional ImageNet Qualitative Examples from Non-PASCAL (unseen) Categories} \\
      \hline
    \includegraphics[width=1.8\columnwidth]{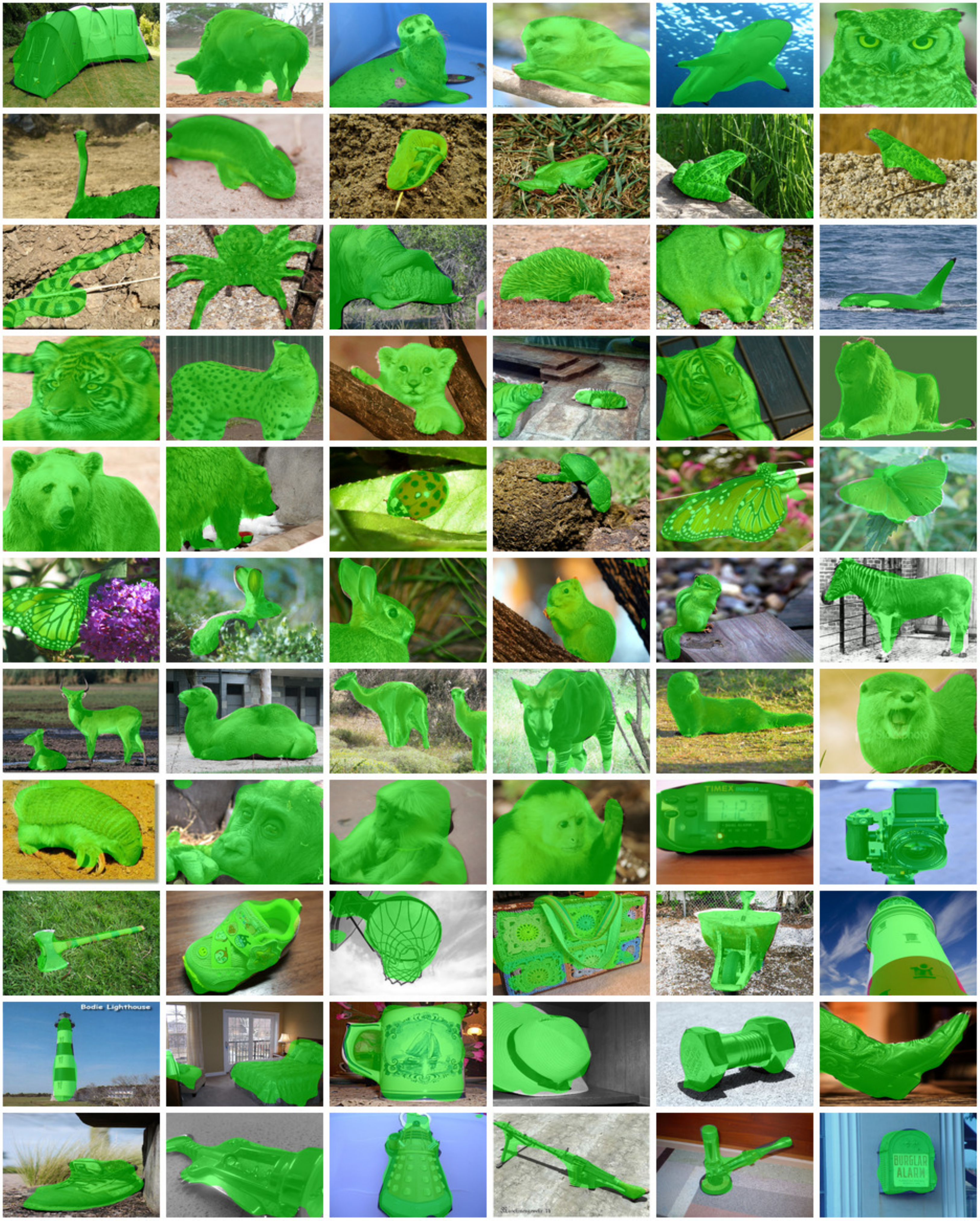} \\
	\hline
	\end{tabular}
	\caption{Qualitative results: We show example segmentations from ImageNet dataset obtained by our pixel objectness model on Non-PASCAL Categories. The segmentation results are shown with a green overlay. Our method generalizes remarkably well and is able to accurately segment foreground objects even for those categories which were never seen during training. Best viewed in color.}
	  \label{fig:qual_res_unseen1}
\end{figure*}

\begin{figure*}[h!]
  \centering
  \renewcommand{\tabcolsep}{0pt}
  \captionsetup{width=1\textwidth, font={footnotesize}, skip=2pt}
   \begin{tabular}{|c|}
      \hline
      {\bf Additional ImageNet Qualitative Examples from Non-PASCAL (unseen) Categories} \\
      \hline
    \includegraphics[width=1.8\columnwidth]{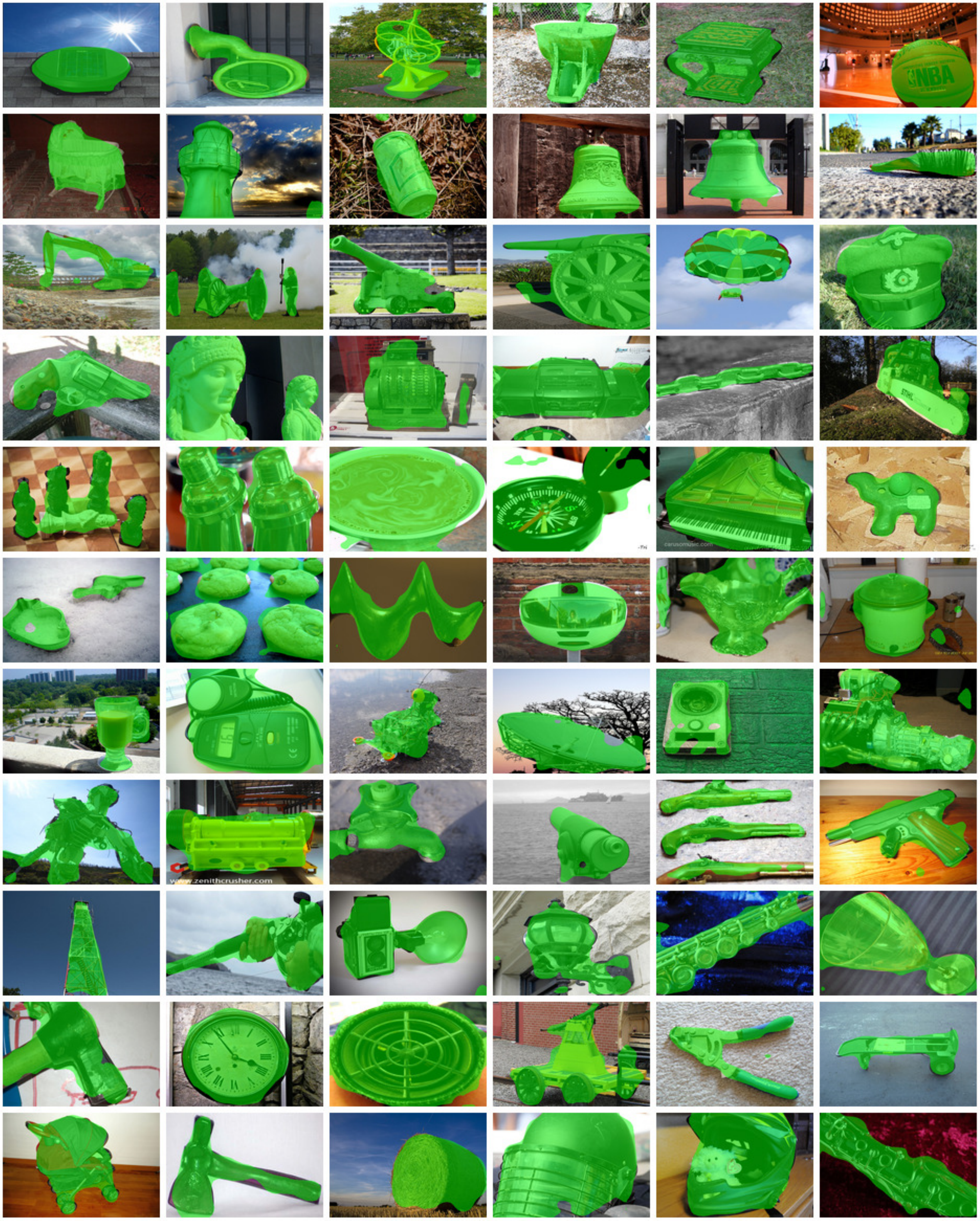} \\
	\hline
	\end{tabular}
	\caption{Qualitative results: We show example segmentations from ImageNet dataset obtained by our pixel objectness model on Non-PASCAL Categories. The segmentation results are shown with a green overlay. Our method generalizes remarkably well and is able to accurately segment foreground objects even for those categories which were never seen during training. Best viewed in color.}
	  \label{fig:qual_res_unseen2}
\end{figure*}

\begin{figure*}[h!]
  \centering
  \renewcommand{\tabcolsep}{0pt}
  \captionsetup{width=1\textwidth, font={footnotesize}, skip=2pt}
   \begin{tabular}{|c|}
      \hline
      {\bf Additional ImageNet Qualitative Examples from Non-PASCAL (unseen) Categories} \\
      \hline
    \includegraphics[width=1.8\columnwidth]{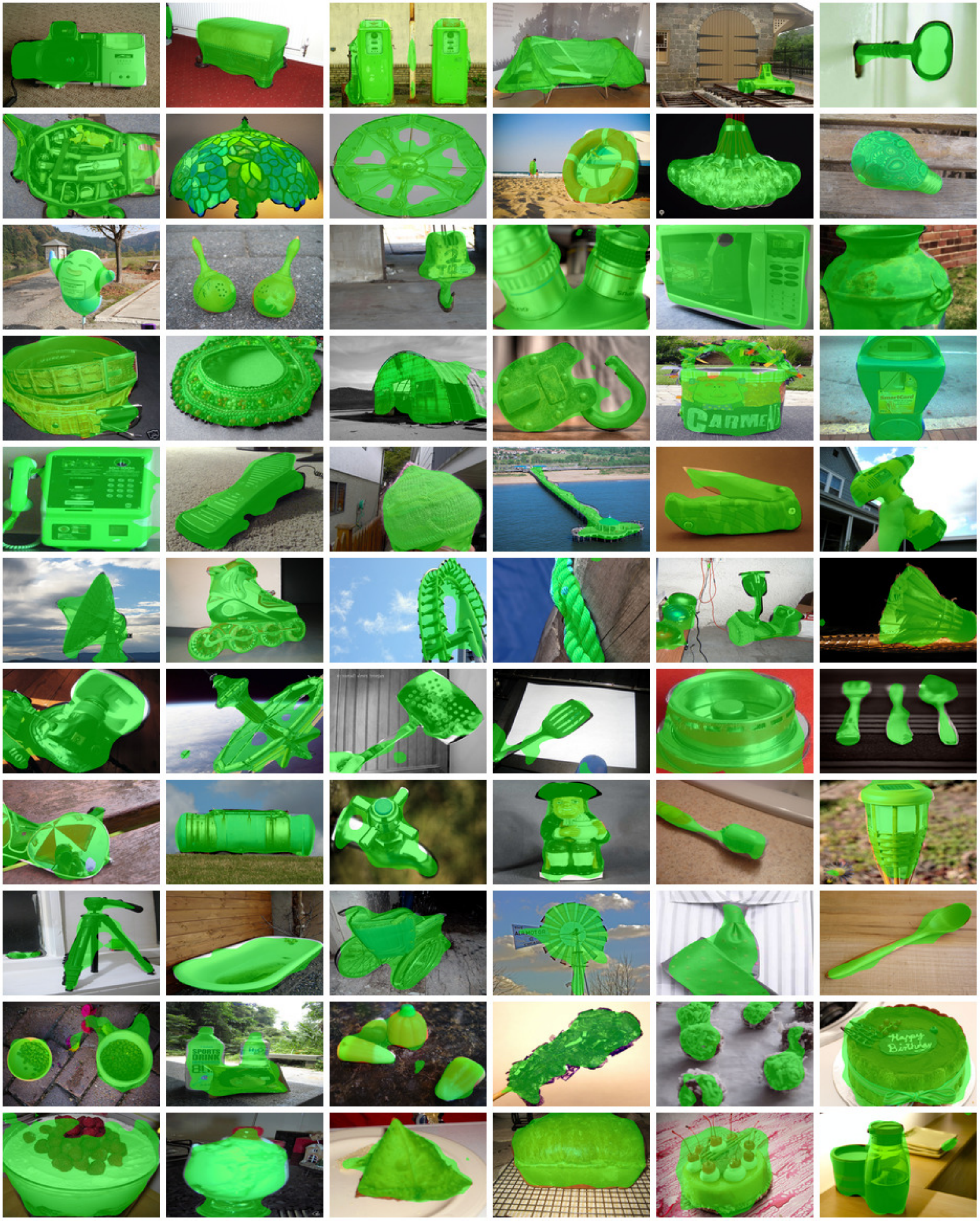} \\
	\hline
	\end{tabular}
	\caption{Qualitative results: We show example segmentations from ImageNet dataset obtained by our pixel objectness model on Non-PASCAL Categories. The segmentation results are shown with a green overlay. Our method generalizes remarkably well and is able to accurately segment foreground objects even for those categories which were never seen during training. Best viewed in color.}
	  \label{fig:qual_res_unseen3}
\end{figure*}

\begin{figure*}[h!]
  \centering
  \renewcommand{\tabcolsep}{0pt}
  \captionsetup{width=1\textwidth, font={footnotesize}, skip=2pt}
   \begin{tabular}{|c|}
      \hline
      {\bf Additional ImageNet Qualitative Examples for Failure Cases} \\
      \hline
    \includegraphics[width=1.8\columnwidth]{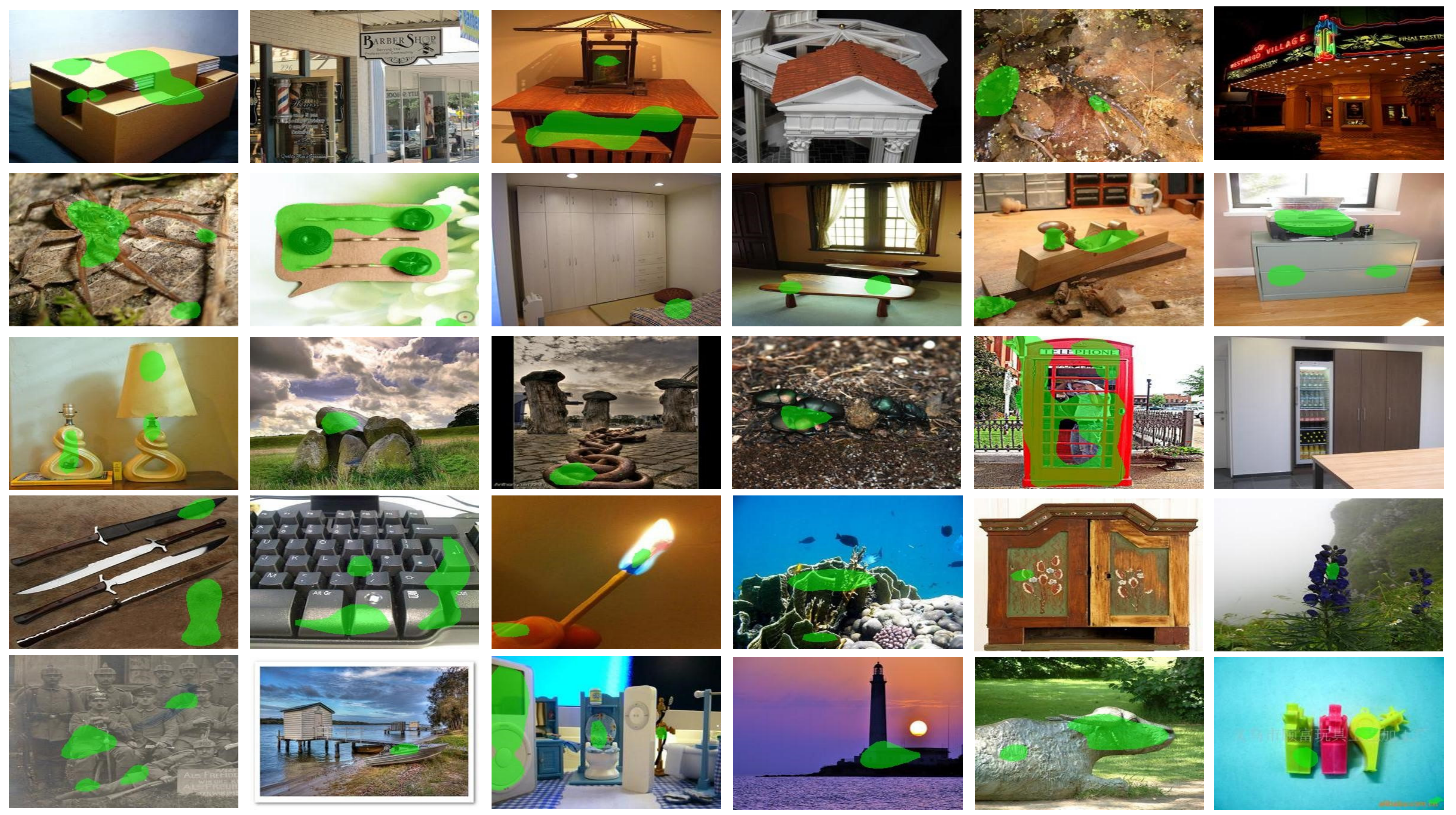} \\
        \hline
        \end{tabular}
        \caption{Qualitative results: We show examples of failure cases from ImageNet dataset obtained by our pixel objectness model. The segmentation results are shown with a green overlay. Typical failure cases involve scene-centric images or images containing very thin objects. Best viewed in color.}
          \label{fig:fail}
\end{figure*}

Figure~\ref{fig:qual_res_unseen1},~\ref{fig:qual_res_unseen2},~\ref{fig:qual_res_unseen3} show qualitative results for those ImageNet images which belong to the non-PASCAL categories. Even though trained only on foregrounds from PASCAL categories, our method generalizes surprisingly well. As can be seen, it can accurately segment foreground objects from completely disjoint categories, examples of which were never seen during training. Figure~\ref{fig:fail} shows more failure cases.

\subsection{Foreground-aware image retargeting examples (Sec.~\ref{sec:results_mturk} in the main text)}

\begin{figure*}[t]
\centering
\includegraphics[width=1.8\columnwidth]{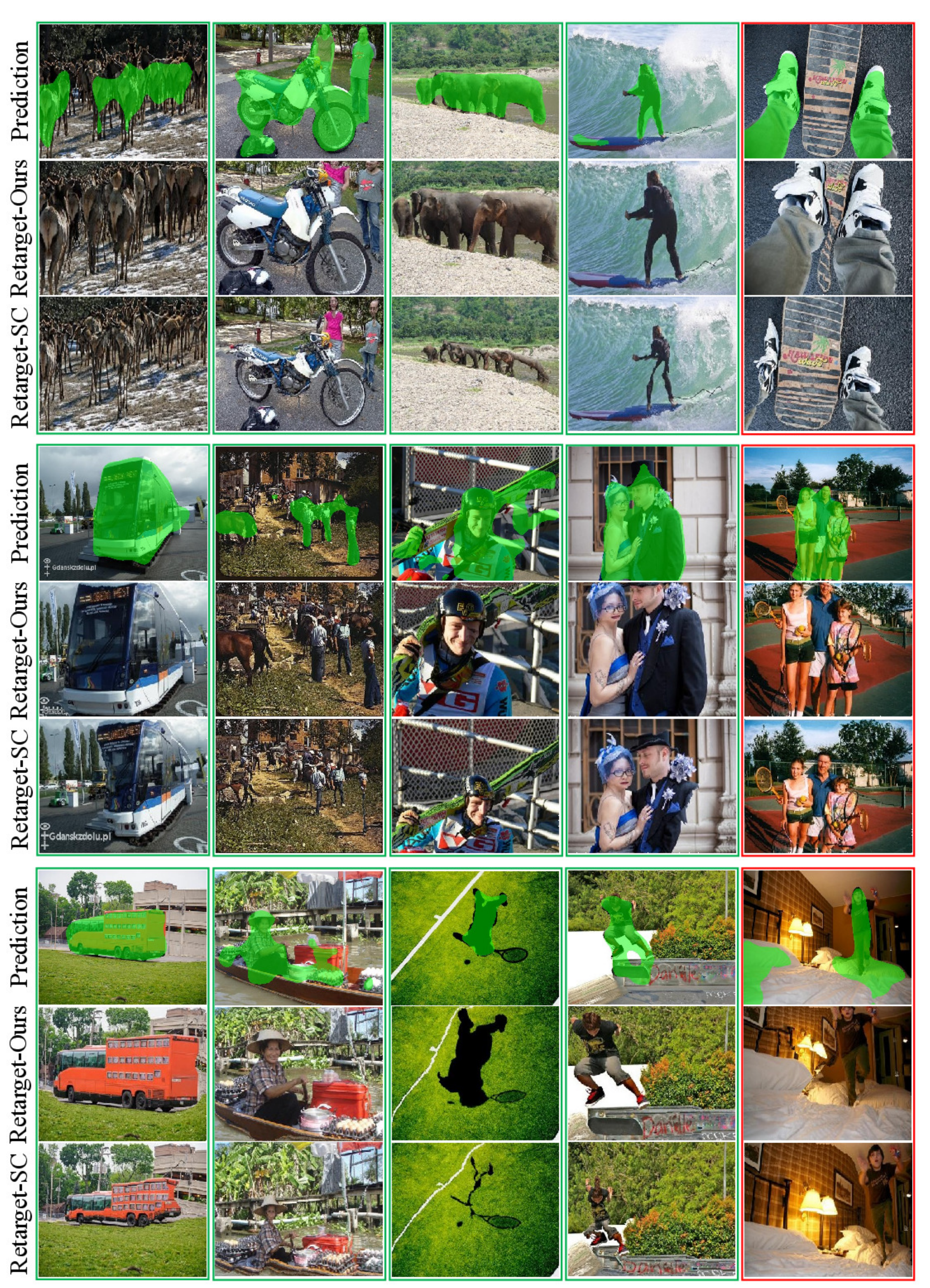}
\caption{We show more foreground-aware image retargeting example results. We show original images with predicted foreground in green (prediction, top row), retargeting images produced by our method (Retarget-Ours, middle row) and retargeting images produced by the Seam Carving based on gradient energy~\cite{avidan2007seam} (Retarget-SC, bottom row). Our method successfully preserves the important visual content while reducing the content of the background. We also present a few failure cases in the rightmost column. Best viewed in color.}
\label{fig:seam_example_supp}
\end{figure*}

We also present more foreground-aware image retargeting example results in Figure~\ref{fig:seam_example_supp}. Please refer to Sec.~\ref{sec:results_mturk} in the main paper for algorithmic details. Our method is able to preserve important objects in the images thanks to the proposed foreground segmentation method. The baseline produces images with important objects distorted, because gradient strength is not a good indicator for perceived content, especially when background is textured. We also present a few failure cases in the rightmost column. In the first example, our method is unsuccessful at predicting the skateboard as the foreground and therefore results in an image with skateboard distorted. In the second example, our method is able to detect and preserve all the people in the image. However, the background distortion creates artifacts that make the resulting image unpleasant to look at compared to the baseline. In the last example, our method misclassified the pillow as foreground and results in an image with an amplified pillow.

\begin{figure*}[t]
\centering
\includegraphics[width=1.8\columnwidth]{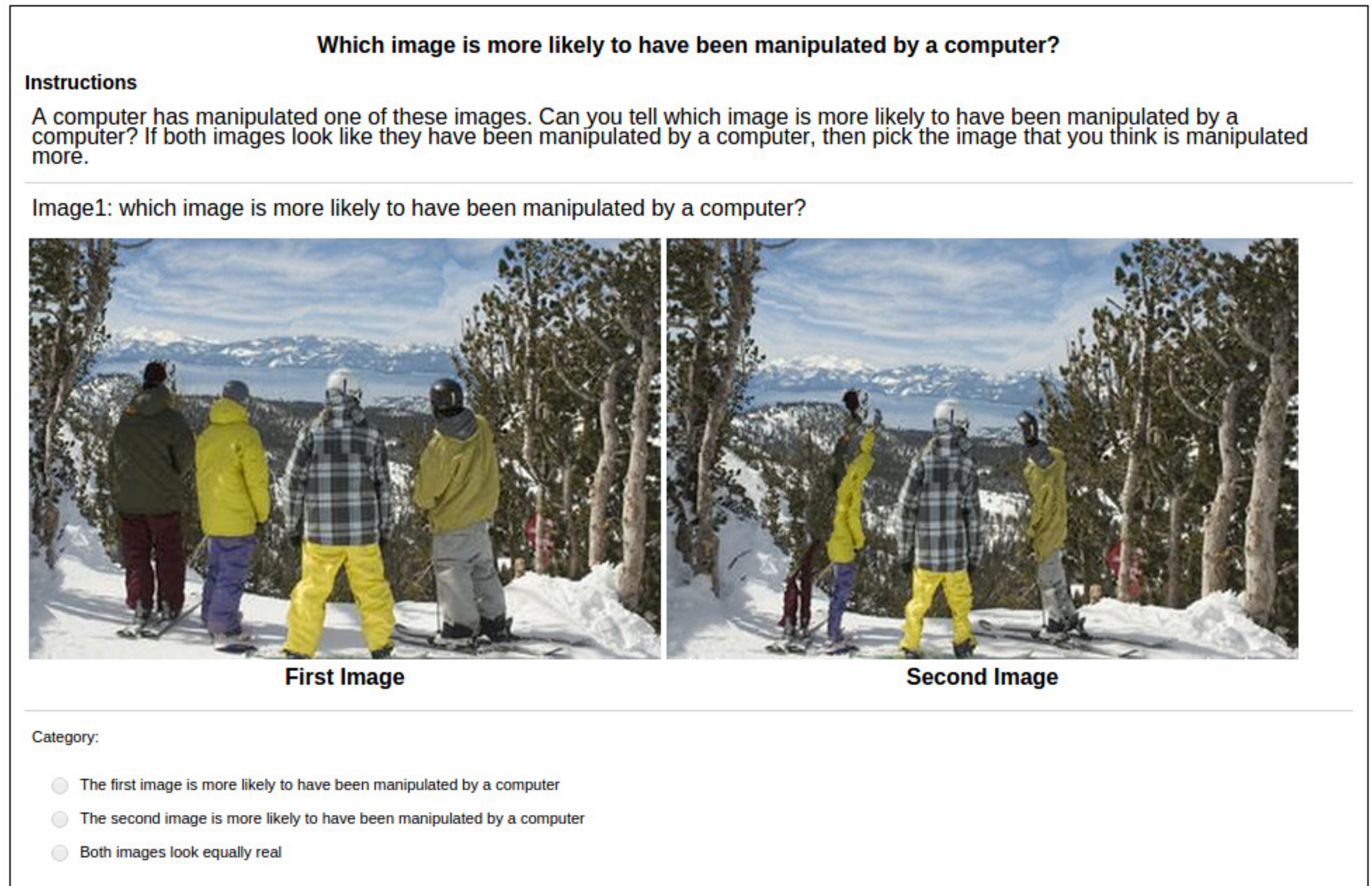}
\caption{Amazon Mechanical Turk interface used to collect human judgement for image retargeting. We ask workers to judge which image is more likely to have been manipulated by a computer. They have three options: 1) The first image is more likely to have been manipulated by a computer; 2) The second image is more likely to have been manipulated by a computer; 3) Both images look equally real.
}
\label{fig:amt_interface}
\end{figure*}
\subsection{Amazon Mechanical Turk interface (Sec.~\ref{sec:results_mturk} in the main text)} We also show the interface we used to collect human judgement for image retargeting on Amazon Mechanical Turk. The two images produced by our algorithm and the baseline method are shown in arbitrary order to the workers. We instruct the workers to pick an image that is more likely to have been manipulated by a computer. If both images look like they have been manipulated by a computer, then pick the one that is manipulated more. The workers have three options to choose from: 1) The first image is more likely to have been manipulated by a computer; 2) The second image is more likely to have been manipulated by a computer; 3) Both images look equally real. See Figure~\ref{fig:amt_interface} for the interface. Also refer to Sec.~\ref{sec:results_mturk} in the main paper for more discussions on these user study results.

\end{document}